\documentclass[journal]{IEEEtran}
\usepackage{cite}
\usepackage{amsmath,amssymb,amsfonts}
\usepackage{algorithmic}
\usepackage{graphicx}
\usepackage{textcomp}
\usepackage{booktabs}
\usepackage{multirow}
\usepackage{array}
\usepackage{url}
\usepackage[hidelinks]{hyperref}
\usepackage{bm}
\usepackage{xcolor}

\newif\ifshowrevisions
\showrevisionsfalse 

\definecolor{revblue}{RGB}{0,82,204}

\newcommand{\rev}[1]{%
    \ifshowrevisions
        \textcolor{revblue}{#1}%
    \else
        #1%
    \fi
}

\def\BibTeX{{\rm B\kern-.05em{\sc i\kern-.025em b}\kern-.08em
    T\kern-.1667em\lower.7ex\hbox{E}\kern-.125emX}}

\begin{document}

\title{TinyNeRV: Compact Neural Video Representations via Capacity Scaling, Distillation, and Low-Precision Inference}

\author{Muhammad Hannan Akhtar,~\IEEEmembership{Graduate Student Member,~IEEE},
        Ihab Amer,~\IEEEmembership{Senior Member,~IEEE},
        and Tamer Shanableh,~\IEEEmembership{Senior Member,~IEEE}%
\thanks{The authors are with the Department of Computer Science and Engineering, American University of Sharjah, Sharjah, United Arab Emirates (e-mail: b00101092@aus.edu; iamer@aus.edu; tshanableh@aus.edu).}%
\thanks{Corresponding author: Muhammad Hannan Akhtar (e-mail: b00101092@aus.edu).}%
\thanks{This work was supported in part by the Open Access Program of the American University of Sharjah, UAE.}%
\thanks{Published in \emph{IEEE Access}, 2026. DOI: \href{https://doi.org/10.1109/ACCESS.2026.3735991}{10.1109/ACCESS.2026.3735991}.}%
}

\markboth{IEEE Access, 2026}%
{Akhtar \emph{et al.}: TinyNeRV: Compact Neural Video Representations}

\maketitle

\begin{abstract}
Neural Representations for Videos (NeRV) encode an entire video sequence within the parameters of a neural decoder, enabling direct frame reconstruction from a temporal index. Although NeRV-style models provide an attractive alternative to conventional sequential decoding pipelines, their deployment in resource-constrained settings remains challenging because reconstruction quality, model capacity, decoding cost, and numerical precision are tightly coupled. Recent efficient NeRV variants primarily reduce complexity through operator-level redesign or selective numerical approximation, while the extreme low-capacity deployment regime remains insufficiently characterized. \rev{This paper presents TinyNeRV, a deployment-oriented design and evaluation framework for compact neural video representations based on controlled capacity scaling, teacher-guided quality recovery, and low-precision inference.} Two lightweight configurations, NeRV-T and NeRV-T+, are introduced to represent minimum-compute and balanced low-capacity operating points. NeRV-T uses only 0.80M parameters and 22.62 GFLOPs, while NeRV-T+ uses 1.68M parameters and 89.80 GFLOPs. To compensate for capacity-induced detail loss without increasing inference cost, we propose frequency--focal knowledge distillation, which emphasizes high-frequency teacher--student discrepancies. This improves PSNR by 0.84 dB for NeRV-T and 0.88 dB for NeRV-T+ on Big Buck Bunny under full-precision inference. We further analyze weights-only quantization under INT8, INT6, and INT4 precision using post-training quantization, quantization-aware training, and distillation-assisted quantization-aware training. \rev{Across the complete seven-sequence UVG benchmark, frequency--focal distillation improves baseline PSNR by an average of 0.70~dB for NeRV-T and 0.78~dB for NeRV-T+. Across UVG and BBB, INT8 remains nearly lossless, INT6 exhibits modest PTQ degradation that is largely recovered by QAT, and INT4 requires retraining-aware adaptation, with teacher supervision providing an additional average gain.} These findings identify practical capacity--supervision--precision trade-offs for deploying neural video representations in real-time and resource-constrained environments.
\end{abstract}


\section{Introduction}
\label{sec:introduction}
\IEEEPARstart{I}{mplicit} neural representations have emerged as a powerful paradigm for modeling visual signals, enabling images, videos, and scenes to be encoded directly within the parameters of compact neural networks. In the context of video, Neural Representations for Video (NeRV)~\cite{chen2021nerv} replace traditional frame-based decoding pipelines with a lightweight decoder that maps a temporal index to a full-resolution frame. This design eliminates sequential frame dependencies and enables frame-independent random-access decoding, offering a compelling alternative to conventional learned video compression systems in latency-sensitive environments.

Recent advances in TinyML and edge intelligence have highlighted the growing importance of compact, efficient neural models capable of operating under strict resource constraints~\cite{roth2024efficient, le2026efftinyml, torre2022iot, yahyati2025systematic}. Across domains such as embedded sensing~\cite{nemeth2025machine, reis2026low, kadhum2025ultra}, on-device inference~\cite{kim2023ondevice, kim2023device}, real-time streaming~\cite{zheng2023streamnet, ye2025edgestreaming}, and edge computing~\cite{bhushan2025deploying, tu2025distributed}, model feasibility is governed not only by reconstruction quality but also by computational complexity, memory footprint, energy consumption, and numerical precision limitations. In these scenarios, aggressively scaled-down models are often required to meet throughput and memory budgets~\cite{berthelier2021deep, liu2024lightweight, kulkarni2024tinyml, heydari2025tiny}. While NeRV-style architectures naturally lend themselves to compact representations, the low-capacity regime (where models are deliberately minimized to maximize efficiency) remains insufficiently explored.

\rev{The practical significance of this low-capacity regime is reflected in the resource budgets of contemporary embedded and edge-AI platforms. Representative hardware spans conventional microcontrollers such as the STM32H753, which provides 1~MB of RAM and 2~MB of flash memory, AI-enabled microcontrollers such as the STM32N657, which integrates 4.2~MB of SRAM and a Neural-ART accelerator rated at up to 600~GOPS, and higher-capacity edge-AI systems such as the NVIDIA Jetson Orin Nano Super, which provides 8~GB of LPDDR5 memory and up to 67 INT8 TOPS within configurable 7--25~W power modes~\cite{stm32h753_datasheet,stm32n6_datasheet,jetson_orin_nano_datasheet,jetson_orin_nano_super_specs}. These hardware tiers motivate reducing both the stored model footprint and per-frame computational cost, and Section~IV-B relates the proposed TinyNeRV operating points directly to these representative resource constraints.}

Existing NeRV variants such as NeRV-S, NeRV-M, and NeRV-L progressively improve reconstruction quality at the cost of increased parameter count, computational load, and reduced decoding speed. However, the design space of \emph{tiny} NeRV models has not been systematically characterized. In particular, it remains unclear how aggressively scaled variants behave across diverse video content, how much quality is sacrificed for efficiency, and whether training-time or deployment-time strategies can compensate for the degradation inherent to extreme model compression.

In this work, we present a systematic study of tiny NeRV variants designed explicitly for constrained deployment. We introduce two lightweight configurations. NeRV-T represents an extreme throughput-oriented operating point optimized for maximal decoding speed and minimal computational cost, while NeRV-T+ provides a carefully balanced design that modestly increases capacity in order to improve reconstruction quality while preserving high efficiency. Together, these models enable structured exploration of the low-capacity regime and allow us to quantify efficiency--quality trade-offs under realistic deployment metrics.

\rev{We evaluate the proposed models on all seven UVG sequences---Beauty, Bosphorus, HoneyBee, Jockey, ReadySetGo, ShakeNDry, and YachtRide---together with Big Buck Bunny (BBB), yielding eight videos in total. Every sequence uses the same 132-frame, $720\times1280$ setting and the same model definitions, sampling, optimization, distillation, and quantization protocols, with no sequence-specific hyperparameter tuning. This controlled benchmark provides a consistent assessment of capacity, distillation, temporal stability, and low-precision behavior across heterogeneous content.}

Beyond architectural scaling, we investigate strategies for mitigating the quality drop observed in tiny models without altering inference complexity. At training time, we explore knowledge distillation (KD) using a higher-capacity teacher model to transfer supervisory signals to compact students. While KD has been widely adopted in classification and generative tasks, its application to neural video representations (and particularly to the tiny regime) has not been systematically studied. At deployment time, we analyze robustness under reduced numerical precision, evaluating INT8, INT6, and INT4 weights-only quantization under quantization-aware training (QAT). We further examine the interaction between distillation and aggressive low-bit inference, revealing capacity-dependent behaviors under severe precision constraints.

\rev{Our experimental results show consistent capacity and training trends across the complete UVG benchmark. INT8 weights are nearly lossless, INT6 introduces moderate PTQ degradation that is largely recovered by QAT, and INT4 exhibits substantial PTQ degradation that requires retraining-aware adaptation. Frequency--focal distillation improves full-precision tiny models on every UVG sequence, while KD-assisted QAT provides a further average improvement at INT4.}

\rev{The novelty of TinyNeRV lies not in introducing a new decoder topology, but in formulating and experimentally characterizing a deployment-oriented low-capacity regime for NeRV models. Existing efficient NeRV variants predominantly improve efficiency through architectural redesign, specialized operators, or numerical approximation of otherwise conventional-capacity models. In contrast, TinyNeRV preserves the original NeRV decoding structure and studies model capacity, training-time supervision, and numerical precision as three coupled deployment axes. Methodologically, the key new component is frequency--focal knowledge distillation, which combines multi-scale low/high-frequency decomposition with a disagreement-derived spatial weighting map so that supervision is concentrated on high-frequency information that is disproportionately lost under severe capacity reduction. This is complemented by a systematic analysis of how the resulting tiny models behave under INT8, INT6, and INT4 weights-only quantization, including the interaction between teacher supervision and quantization-aware training. The practical significance of this formulation is that the resulting operating points correspond to concrete reductions in stored-weight footprint and per-frame computation, allowing the models to cross memory and compute thresholds relevant to embedded and edge-AI platforms rather than providing only relative compression ratios.}

The main contributions of this paper are summarized as follows:

\begin{itemize}
    \item \rev{We establish and systematically characterize the extreme low-capacity regime of NeRV-style neural video representations, defining NeRV-T and NeRV-T+ as controlled deployment operating points that preserve the original decoding topology while substantially reducing stored-weight footprint and per-frame computation.}

    \item \rev{We propose frequency--focal knowledge distillation for capacity-constrained neural video decoders. Unlike uniform output distillation, the proposed objective combines multi-scale frequency decomposition with teacher--student disagreement weighting to selectively emphasize high-frequency detail that is most vulnerable to severe width reduction, without altering inference-time complexity.}

    \item \rev{We validate the resulting capacity and training trends across the complete seven-sequence UVG benchmark plus BBB and jointly study INT8, INT6, and INT4 weights-only quantization under PTQ, QAT, and KD-assisted QAT. This reveals distinct capacity--precision regimes and shows when teacher supervision remains beneficial under aggressive numerical compression.}

    \item \rev{We relate these model operating points to representative embedded and edge-AI resource constraints and derive deployment-oriented guidelines for selecting between extreme-throughput, balanced-quality, and low-bit configurations under memory, compute, and reconstruction-quality requirements.}
\end{itemize}

Together, these findings provide a technically rigorous and deployment-focused understanding of tiny neural video representations, bridging the gap between architectural design and real-world feasibility in constrained systems.

\section{Related Work}

Neural video representations transform video signals into parameterized models learned by neural networks. 
This paradigm encompasses implicit neural representations (INRs), image-wise decoder frameworks such as NeRV and its derivatives, and a growing line of work that addresses efficiency, expressiveness, and deployment constraints. 
The following subsections review the literature on neural video representations, model compression and capacity scaling, knowledge distillation in generative and regression models, and low-precision/low-bit inference techniques relevant to deployment in constrained environments. 
Our review is structured to situate the proposed TinyNeRV framework within the broader research landscape, highlight key methodological advances, and identify gaps in existing work that motivate the current study.

\subsection{Neural Video Representations}

Implicit neural representations (INRs) model visual signals as continuous functions parameterized by neural networks. 
Instead of storing pixels explicitly, INRs learn a mapping $\mathcal{F}_\theta : (x,y,t) \rightarrow \mathbf{RGB}$, where the network parameters $\theta$ implicitly encode the signal. 
Early coordinate-based approaches such as SIREN~\cite{sitzmann2020siren} demonstrated that periodic activation functions enable accurate modeling of high-frequency signals. 
However, coordinate-wise decoding requires evaluating the network at every spatial coordinate, resulting in significant computational overhead for high-resolution video reconstruction.

To address this limitation, Chen \emph{et al.} introduced Neural Representations for Videos (NeRV)~\cite{chen2021nerv}, which reformulated video INR as an image-wise generation task. 
Rather than predicting RGB values per coordinate, NeRV reconstructs an entire frame directly from a temporal index using a convolutional decoder with hierarchical upsampling blocks. 
This formulation enables frame-wise random-access decoding without sequential dependencies and significantly reduces sampling complexity.

Subsequent works expanded the NeRV framework along architectural and representational dimensions. 
HNeRV~\cite{li2023hnerv} introduced hybrid learned embeddings to enhance representational flexibility and improve internal generalization across frames. 
HiNeRV~\cite{hinerv2023} proposed hierarchical encoding strategies to better allocate capacity across spatial resolutions, improving modeling of fine structures and large-scale coherence. 
Res-NeRV~\cite{tarchouli2024resnerv} incorporated residual connections into the decoder to improve gradient flow and representation stability within the image-wise decoding paradigm.

Another important line of work addresses spectral bias in neural representations. 
Neural networks tend to learn low-frequency components more rapidly than high-frequency details, which can result in over-smoothed reconstructions. 
High-frequency enhanced hybrid representations~\cite{highfreq2023} and spatio-temporal spectra-preserving approaches~\cite{kim2026snervplus} incorporate frequency decomposition or spectral regularization mechanisms to improve detail preservation and temporal consistency. 
These methods aim to explicitly model multi-scale spatial and temporal frequency components within the implicit representation.

Collectively, neural video representation research has evolved from coordinate-based INR formulations toward structured image-wise decoders with enhanced embedding strategies and frequency-aware modeling. 
However, most prior work emphasizes improving representational fidelity through architectural enrichment rather than explicitly studying extreme low-capacity regimes under strict deployment constraints.

\subsection{Model Compression and Capacity Scaling}

In neural video representations, the learned network parameters themselves constitute the encoded form of the video, directly coupling representation fidelity with storage footprint and computational cost. Unlike conventional codecs where bitrate and decoding complexity are partially decoupled, INR-based video modeling intrinsically links reconstruction quality to model capacity. Broader literature on neural network scaling demonstrates that performance is governed by structured allocation of representational resources across depth, width, and resolution~\cite{jia2025towards}. Compound scaling strategies, such as EfficientNet~\cite{tan2019efficientnet}, show that balanced scaling yields more favorable accuracy--efficiency trade-offs than naive expansion along a single dimension. In convolutional architectures, width scaling in particular provides predictable control over parameter count and multiply--accumulate operations, making it a practical mechanism for regulating computational complexity.

Model compression techniques further seek to reduce parameter count while preserving performance. Structured pruning removes entire channels or filters to produce hardware-friendly reductions in computation, whereas unstructured pruning targets individual weights to minimize storage overhead. Quantization provides an orthogonal compression axis by reducing numerical precision, thereby decreasing memory footprint and enabling integer-arithmetic inference. Post-training quantization (PTQ) discretizes weights after optimization, while quantization-aware training (QAT) incorporates simulated quantization during training to improve robustness under low-bit representations~\cite{jacob2018quantization, bahri2024neural}. Although such techniques are well established in classification networks, generative and reconstruction-based models tend to exhibit greater sensitivity to aggressive compression due to their direct pixel-level objectives and limited tolerance for approximation error~\cite{sengupta2025how, shao2025memory}.

Recent efficient NeRV variants have also explored deployment-oriented compression from architectural and numerical perspectives. Binary-NeRV~\cite{shanableh2026binary} reduces storage and equivalent computation by selectively replacing convolutional layers with hybrid-precision binary operations, showing that aggressive numerical simplification can improve efficiency but must be applied carefully to avoid severe reconstruction degradation. Ghost-NeRV~\cite{shanableh2026ghost} instead targets operator-level redundancy by replacing standard convolutions with Ghost convolutions, reducing the computational cost of NeRV-style decoding while preserving full-precision inference. These works demonstrate that efficient neural video representation can be pursued through layerwise numerical approximation or convolutional operator redesign.

Within learned compression frameworks, rate--distortion theory formalizes the trade-off between representation size and reconstruction accuracy. In model-based video representation, this trade-off manifests as a relationship between parameter count, effective bitrate, and reconstruction error. However, systematic characterization of extreme low-capacity regimes, where channel width is aggressively reduced while architectural depth and decoding structure remain fixed, has received limited attention~\cite{kwan2025ultra}. Furthermore, interactions between architectural scaling and reduced numerical precision may produce nonlinear degradation when representational redundancy is minimal~\cite{cordova2025edge}. Understanding these coupled effects is essential for deployment-oriented neural video systems operating under strict compute and memory constraints. In contrast to prior work that primarily investigates operator-level redesign, selective precision reduction, moderate scaling, or post-hoc compression of larger models, this study focuses explicitly on controlled width-based scaling and jointly analyzes architectural capacity, knowledge distillation, and low-precision inference in the tiny regime.

\subsection{Knowledge Distillation for Compact Neural Models}

Knowledge distillation (KD) is a widely used technique for transferring information from a high-capacity teacher model to a smaller student model. Originally introduced by Hinton \emph{et al.}~\cite{hinton2015distilling}, KD leverages soft predictions produced by the teacher network to provide richer supervisory signals than ground-truth labels alone. Subsequent work extended this framework beyond output-level supervision to intermediate feature representations and relational structures. For example, FitNets~\cite{romero2014fitnets} proposed transferring intermediate feature activations to guide training of thinner networks, while attention transfer~\cite{zagoruyko2017paying} encouraged students to mimic spatial attention maps of the teacher. These methods demonstrate that structured supervision can significantly improve performance of compact models by preserving informative representations learned by larger networks~\cite{moslemi2024kdsurvey, mansourian2025kdsurvey}.

Knowledge distillation has since been applied across a wide range of computer vision tasks, including image classification, object detection, and semantic segmentation~\cite{wang2022kdvisual, li2026kdvisual, aalishah2025kdvisual, li2025kdvisual, wang2025kdvisual, himeur2025kdvisual}. In addition to feature-based distillation, recent studies explore relational and contrastive distillation strategies that align pairwise sample relationships or representation geometry between teacher and student networks~\cite{Shao_2025_ICCV, zhu2025contrastivekd, Xu_2025_WACV}. Such approaches provide stronger regularization for compact models and often improve training stability when representational capacity is limited ~\cite{chen2026gated, Zheng_Wang_Yuan_2022, wu2024aligning}. While KD is extensively studied in discriminative settings ~\cite{jimaging10040085, xiao2025har}, its application to generative and reconstruction-oriented models is comparatively less explored~\cite{Zhang_2024_CVPR}.

For models that directly predict continuous signals, such as image restoration or super-resolution networks, distillation has been shown to improve reconstruction fidelity by transferring fine-grained spatial or frequency information from a larger teacher model~\cite{Zhao_2025_CVPR, drones9030209, dharejo2024superres}. Frequency-aware distillation strategies further emphasize alignment of high-frequency components that are often difficult for compact networks to learn ~\cite{li2026knowledge, Li_2025_ICCV}. Despite these advances, knowledge distillation has rarely been investigated in the context of neural video representations, where the student model must reconstruct entire frames from a compact neural encoding. In such settings, representational bandwidth is severely constrained, and the ability to transfer structural information from larger teacher models may play a critical role in maintaining reconstruction quality~\cite{kim2025snerv}. 
Motivated by this gap, this work investigates knowledge distillation as a training-time strategy for improving the performance of extremely compact NeRV variants.

\rev{More broadly, recent studies of teacher--student learning indicate that knowledge transfer is not uniform across representational structures and can depend strongly on student capacity and the type of information being distilled. For example, recent work on tabular anomaly detection shows that compact students can preserve global teacher behavior substantially better than local or neighborhood-dependent anomaly structure, highlighting that some forms of knowledge are inherently more difficult to transfer under compression~\cite{krcmar2026tabularKD}. Related studies similarly emphasize that large teacher--student capacity gaps can limit transfer effectiveness and motivate student-oriented or adaptive distillation objectives~\cite{shen2024sokd,tambat2025mcdistil}. These observations are consistent with the motivation of frequency--focal distillation: rather than requiring a capacity-limited student to reproduce all teacher information uniformly, the proposed objective allocates greater supervision to spatial and frequency components that are most vulnerable to compression.}

\rev{Existing frequency-aware and focal distillation methods motivate related principles, but differ from the proposed formulation in both construction and target setting. Wavelet knowledge distillation decomposes generated images into wavelet bands and distills the high-frequency components for image-to-image translation~\cite{zhang2022waveletKD}, while frequency-attention distillation introduces a learnable frequency-domain filter for intermediate feature transfer~\cite{pham2024frequencyKD}. Focal and Global Distillation instead emphasizes informative spatial regions and channels in detector feature maps without explicitly separating structural and high-frequency reconstruction content~\cite{yang2022focalglobalKD}. More recently, FreeKD combines frequency-domain representations with a learned frequency prompt and pixel-wise masking for dense prediction~\cite{zhang2024freekd}. In contrast, frequency--focal distillation is designed as a purely output-level objective for capacity-constrained neural video decoders: multi-scale Gaussian decomposition determines \emph{what} information is transferred by separating low-frequency structure from high-frequency residual detail, while a detached teacher--student RGB disagreement map determines \emph{where} the high-frequency supervision is emphasized. No learned attention module, feature adapter, or additional inference-time component is introduced.}

\subsection{Low-Precision and Quantized Neural Inference}

Low-precision inference is a standard approach for reducing model storage, memory bandwidth, and compute cost in resource-constrained deployment~\cite{abushahla2026neuralnetworkquantizationmicrocontrollers}. 
Quantization replaces full-precision weights (and optionally activations) with low-bit representations, enabling efficient integer arithmetic and substantially reducing the footprint of neural models~\cite{liu2025lowbitmodelquantizationdeep, Yang_2019_CVPR}. 
A widely adopted formulation is uniform affine quantization, where floating-point weights are mapped to discrete levels using a scale and zero-point, typically calibrated per tensor or per channel. 
Early integer-only inference pipelines demonstrated that quantized convolutional networks can retain high accuracy while executing efficiently on commodity hardware~\cite{jacob2018quantization}.

Two primary deployment paradigms are commonly considered. 
PTQ applies discretization after training using calibration data, providing a simple and low-cost path to deployment. 
However, PTQ becomes increasingly brittle at lower bit-widths as quantization noise grows and model redundancy decreases. 
QAT addresses this limitation by simulating quantization effects during training and optimizing weights to be robust to discretization. 
Modern QAT methods typically rely on straight-through estimators to propagate gradients through non-differentiable rounding operations~\cite{yin2018understanding, schoenbauer2025custom, pmlr-v162-nagel22a}, and they are often required when targeting 4-bit or lower precision regimes~\cite{lee2025littlebit, chen2026stableqat, tabesh2025cage}.

Recent work has pushed quantization to aggressive low-bit settings, including 4-bit and mixed-precision inference. 
Approaches such as learned step-size quantization (LSQ)~\cite{esser2020lsq} optimize quantization scales jointly with network weights to better preserve accuracy under low-bit constraints, while post-training methods such as data-free or calibration-based techniques have been explored to reduce the need for retraining~\cite{Bai_2023_ICCV, limin2024quantizationsurvey}. 
In classification networks, these methods can maintain strong performance even under severe quantization, particularly when per-channel quantization and carefully designed calibration procedures are used~\cite{rokh2023quantizationclassification, cho2020perchannel}.

In reconstruction-oriented and generative settings, low-precision inference is often more challenging~\cite{li2025texture, hong2022cadyq}. 
Unlike classification, where small perturbations may not change the predicted label, pixel-level regression requires preserving fine-grained numerical relationships that directly affect visual fidelity~\cite{qin2025visualcomm, deng2023pixelregression}. 
Quantization error can therefore manifest as structured artifacts, loss of high-frequency detail, or amplified reconstruction noise, and these effects can be more pronounced in compact models with limited redundancy~\cite{seo2022quantcnn}. 
In neural video representations, this sensitivity is particularly relevant because the model parameters themselves constitute the encoded video representation. Any distortion introduced by low-precision arithmetic therefore affects both storage efficiency and reconstruction quality simultaneously. Furthermore, compact decoders possess fewer channels and reduced representational redundancy, limiting their ability to absorb quantization-induced perturbations through alternative parameter configurations. As model capacity decreases, maintaining reconstruction fidelity increasingly depends on how effectively the remaining parameters utilize the available numerical precision.
Accordingly, understanding their interaction is essential: aggressive model scaling reduces representational slack, while aggressive quantization injects discretization noise, and the compounded effect may produce nonlinear degradation~\cite{yan2020towards, ouyang2025low}.

Motivated by these considerations, this work evaluates low-precision inference in the tiny NeRV regime using weights-only quantization across INT8, INT6, and INT4. 
By analyzing PTQ versus QAT and studying their interaction with training-time supervision, we provide a deployment-oriented characterization of precision--capacity trade-offs in compact neural video representations.

\rev{Taken together, prior work addresses complementary aspects of neural video representation efficiency, but these directions are largely studied in isolation. Architectural variants such as HNeRV, HiNeRV, and Res-NeRV primarily improve representation fidelity, while Binary-NeRV and Ghost-NeRV reduce deployment cost through numerical approximation and operator-level redesign, respectively. Knowledge-distillation studies show that compact models can recover information from larger teachers, and low-precision inference provides an orthogonal means of reducing storage and computational requirements. However, the interaction between extreme architectural capacity reduction, teacher-guided quality recovery, and aggressive numerical precision has not been systematically characterized for NeRV-style video representations. This gap motivates TinyNeRV, which preserves the underlying NeRV decoding structure and jointly studies controlled width scaling, frequency--focal distillation, and INT8/INT6/INT4 weights-only quantization to characterize practical capacity--supervision--precision trade-offs in the tiny regime.}

\section{Methodology}
\subsection{Neural Video Representation Background}

Neural Video Representation (NeRV)~\cite{chen2021nerv} models a video as a continuous function of time parameterized by a neural network. Instead of treating a video as a sequence of frames, the entire sequence is encoded within the weights of a model that maps a normalized temporal index to its corresponding RGB frame. Video encoding is formulated as fitting a neural network to all frames of a target sequence, while decoding reduces to a single forward pass to retrieve any frame.

\begin{figure*}[!t]
    \centering
    \includegraphics[width=\linewidth]{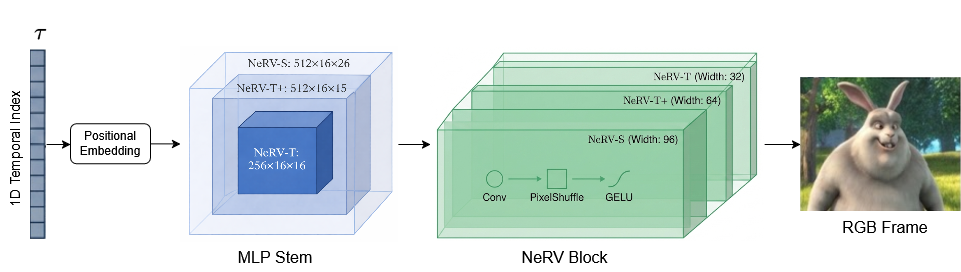}
    \caption{Width-based scaling of NeRV variants. All models share identical depth and decoding structure; capacity differences arise solely from channel dimensionality.}
    \label{fig:tiny_architecture}
\end{figure*}

\rev{Following the original NeRV formulation~\cite{chen2021nerv}, each discrete frame index is first converted to a normalized temporal coordinate. For a video containing $N$ frames, frame $n$ is represented as $\tau_n=n/N$, where $n\in\{0,\ldots,N-1\}$. Rather than supplying this scalar directly to the decoder, we map it to a fixed higher-dimensional sinusoidal representation}
{\ifshowrevisions\color{revblue}\fi
\begin{equation}
\mathrm{PE}(\tau_n)=
\big[
\sin(\pi\tau_n b^i),\cos(\pi\tau_n b^i)
\big]_{i=0}^{L-1}.
\end{equation}
}
\rev{where $b$ determines the geometric spacing of temporal frequencies and $L$ is the number of frequency levels. All experiments use $b=1.25$ and $L=40$, producing an 80-dimensional positional embedding. The use of multiple sinusoidal frequencies exposes the subsequent MLP to temporal variations at different scales, making it easier to represent both slowly varying frame content and more rapid frame-to-frame changes than when operating directly on a single scalar coordinate. The encoded temporal vector is processed by the MLP stem and reshaped into the initial spatial feature map, which is then progressively transformed by the convolutional upsampling blocks into the RGB frame. This positional encoding is fixed across all NeRV capacities considered in this work so that differences between NeRV-T, NeRV-T+, and the larger baselines arise from architectural capacity rather than changes to the temporal representation.}

The model is optimized using a reconstruction objective that combines pixel-wise fidelity and structural similarity. Since the video is fully encapsulated within the model parameters, standard compression techniques such as pruning and quantization can be applied directly to reduce storage and memory requirements.

While prior work demonstrated the feasibility of neural video representations at moderate and large scales, their behavior in the extreme low-capacity regime remains largely unexplored. Aggressive parameter reduction introduces challenges including underfitting, quality degradation, and instability. This work focuses on this constrained setting and systematically investigates how carefully designed tiny variants can retain the structural advantages of NeRV while achieving high throughput and reduced memory requirements.

\subsection{Design of Tiny NeRV Variants}

\begin{table*}[h!] \centering \scriptsize \setlength{\tabcolsep}{3pt} \caption{Layer-wise arithmetic and GFLOPs for NeRV-T, NeRV-T+, and NeRV-S on 720$\times$1280 video (BBB). For the $3\times3$ convolution blocks, multiplications are computed as $\text{Mult} = (HW)\,C_{out}\,(9C_{in})$, and GFLOPs $=2\cdot\text{Mult}/10^9$. PixelShuffle operations involve only tensor rearrangement and incur negligible arithmetic cost.} \label{tab:nerv_flops_breakdown} \resizebox{\textwidth}{!}{%
\begin{tabular}{l c c c c} \toprule \textbf{Stage} & \textbf{Conv grid} & \textbf{NeRV-T} & \textbf{NeRV-T+} & \textbf{NeRV-S} \\ & $(H\times W)$ & Arithmetic $\rightarrow$ GFLOPs & Arithmetic $\rightarrow$ GFLOPs & Arithmetic $\rightarrow$ GFLOPs \\ \midrule Block1 ($s=5$) & $9\times16$ & $144\cdot400\cdot(9\cdot16)\rightarrow0.0166$ & $144\cdot375\cdot(9\cdot15)\rightarrow0.0146$ & $144\cdot650\cdot(9\cdot26)\rightarrow0.0438$ \\ Block2 ($s=2$) & $45\times80$ & $3600\cdot128\cdot(9\cdot16)\rightarrow0.1327$ & $3600\cdot256\cdot(9\cdot15)\rightarrow0.2488$ & $3600\cdot384\cdot(9\cdot26)\rightarrow0.6470$ \\ Block3 ($s=2$) & $90\times160$ & $14400\cdot128\cdot(9\cdot32)\rightarrow1.0617$ & $14400\cdot256\cdot(9\cdot64)\rightarrow4.2467$ & $14400\cdot384\cdot(9\cdot96)\rightarrow9.5551$ \\ Block4 ($s=2$) & $180\times320$ & $57600\cdot128\cdot(9\cdot32)\rightarrow4.2467$ & $57600\cdot256\cdot(9\cdot64)\rightarrow16.9869$ & $57600\cdot384\cdot(9\cdot96)\rightarrow38.2206$ \\ Block5 ($s=2$) & $360\times640$ & $230400\cdot128\cdot(9\cdot32)\rightarrow16.9869$ & $230400\cdot256\cdot(9\cdot64)\rightarrow67.9477$ & $230400\cdot384\cdot(9\cdot96)\rightarrow152.8824$ \\ RGB head ($1\times1$) & $720\times1280$ & $921600\cdot3\cdot32\rightarrow0.1769$ & $921600\cdot3\cdot64\rightarrow0.3539$ & $921600\cdot3\cdot96\rightarrow0.5308$ \\ \midrule \textbf{Total GFLOPs} & & \textbf{22.6216} & \textbf{89.7987} & \textbf{201.8797} \\ \bottomrule \end{tabular}%
} \end{table*}

\rev{TinyNeRV preserves the NeRV decoding depth, stride schedule, kernel sizes, and output resolution, and varies capacity only through channel width. After the initial feature expansion, the channel width entering stage $i+1$ follows}
\begin{equation}
C_{i+1}=\max\left(\left\lfloor\frac{C_i}{r}\right\rfloor,C_{\min}\right),
\end{equation}
\rev{where $r=2$ and $C_{\min}$ is the minimum decoder width. Capacity is therefore controlled through the MLP stem, the $9\times16$ seed-feature width, and the high-resolution width floor $C_{\min}$.}

\rev{NeRV-S uses a 512-channel stem, a 26-channel seed map, and a 96-channel floor. NeRV-T+ retains the 512-channel stem but reduces the seed width to 15 and $C_{\min}$ to 64, whereas NeRV-T uses a 256-channel stem, a 16-channel seed map, and a 32-channel floor. Because the floor is reached in the later stages, NeRV-T and NeRV-T+ remain constrained to 32 and 64 channels, respectively, precisely where the feature maps are largest. NeRV-T therefore prioritizes minimum computation and throughput, while NeRV-T+ allocates additional high-resolution capacity for better reconstruction quality. These two configurations are selected as the primary lightweight operating points because they intentionally span two distinct regions of the low-capacity design space: NeRV-T represents an aggressive minimum-capacity setting, whereas NeRV-T+ restores a modest amount of high-resolution width to provide a more balanced quality--complexity trade-off while remaining substantially smaller than NeRV-S. Studying both variants therefore allows the subsequent distillation and quantization experiments to determine whether the proposed recovery strategies remain effective under both extreme and moderately relaxed capacity constraints. Figure~\ref{fig:tiny_architecture} summarizes this controlled width scaling.}

\rev{Table~\ref{tab:nerv_flops_breakdown} explains why these width changes have a large computational effect. For PixelShuffle factor $s$, the preceding convolution produces $C_{\mathrm{out}}s^2$ channels at $H\times W$, which are rearranged into $C_{\mathrm{out}}$ channels at $sH\times sW$; the number of spatial positions therefore increases by $s^2$. Most stages use $s=2$, so each step quadruples spatial area. PixelShuffle itself is only a rearrangement, but the surrounding convolutions operate on progressively larger maps, making channel reduction in the high-resolution stages the dominant source of GFLOP savings. Since the backbone is unchanged, performance differences across variants can be attributed directly to representational width rather than structural modification.}

\subsection{Knowledge Distillation for Tiny Neural Video Representations}

\begin{figure*}[!h]
    \centering
    \includegraphics[width=\linewidth]{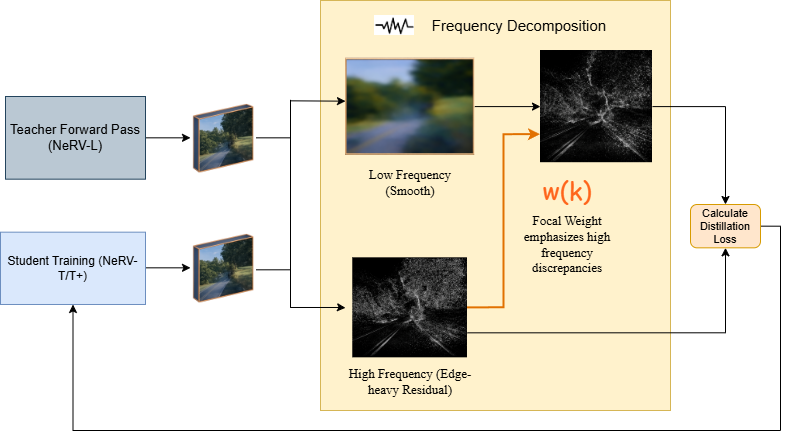}
    \caption{\rev{Proposed frequency--focal knowledge distillation. Teacher and student predictions are decomposed into low-frequency (smooth) and high-frequency (edge-dominant residual) components. A focal weighting map $w_s$, derived from the normalized teacher--student RGB disagreement at each scale, emphasizes difficult spatial locations when weighting the high-frequency distillation term, guiding the tiny student toward improved detail reconstruction while preserving global structure.}}
    \label{fig:kd_ff}
\end{figure*}

\rev{Aggressive width reduction reduces computation but also promotes underfitting, over-smoothing, and loss of fine detail. Rather than recovering quality by widening the high-resolution decoder, we use knowledge distillation (KD), which strengthens training supervision without changing the student's inference-time architecture. To our knowledge, this is the first systematic study of KD for NeRV models in the extreme tiny regime. We select NeRV-L as the teacher because it is the highest-capacity model in the controlled NeRV family evaluated in this work and provides the strongest reconstruction target while retaining the same underlying decoder formulation as the tiny students. Using a teacher from the same architectural family also avoids introducing a separate teacher architecture as a confounding factor, so the experiment isolates the transfer of representational capacity from a larger NeRV model to NeRV-T and NeRV-T+. The pretrained NeRV-L teacher is frozen throughout distillation, while NeRV-T and NeRV-T+ are optimized with ground-truth reconstruction and distillation losses:}
\begin{equation}
\mathcal{L} = \mathcal{L}_{\text{GT}} + \lambda_{\text{KD}}\,\mathcal{L}_{\text{KD}},
\end{equation}
\rev{where $\lambda_{\text{KD}}$ controls teacher guidance. The student otherwise follows the baseline training pipeline. The simplest strategy is final-output KD, which directly matches the student and teacher RGB reconstructions for temporal index $\tau$:}
\begin{equation}
\mathcal{L}_{\text{KD}}^{\text{final}} = \| S(\tau) - T(\tau) \|_1 .
\end{equation}
\rev{Output-level matching avoids feature-width alignment and adds no inference-time components, but it does not explicitly target the fine detail most affected by severe width reduction. We therefore propose frequency--focal distillation, which separates coarse structure from high-frequency detail and then emphasizes detail supervision at locations of larger teacher--student disagreement. Figure~\ref{fig:kd_ff} summarizes the pipeline.}

\rev{Frequency--focal distillation is applied at multiple output scales. Let $S_s$ and $T_s$ denote the student and teacher predictions at scale $s\in\{1,2,4\}$, where $s=1$ corresponds to the original output resolution and the coarser scales are obtained using adaptive average pooling. At each scale, a Gaussian low-pass filter is applied independently to each color channel to obtain the low-frequency component, while the corresponding high-frequency residual is defined by subtraction:}
{\ifshowrevisions\color{revblue}\fi
\begin{equation}
S_{s,\text{low}}=\mathcal{G}_{k,\sigma}(S_s),\qquad
S_{s,\text{high}}=S_s-S_{s,\text{low}},
\end{equation}
\begin{equation}
T_{s,\text{low}}=\mathcal{G}_{k,\sigma}(T_s),\qquad
T_{s,\text{high}}=T_s-T_{s,\text{low}}.
\end{equation}
}

\rev{The low-frequency term encourages the student to match global structure, color distribution, and coarse shading, whereas the high-frequency term targets edges, thin structures, and textural detail that are disproportionately affected by severe width reduction. Without focal weighting, the scale-specific frequency distillation objective is}
{\ifshowrevisions\color{revblue}\fi
\begin{equation}
\mathcal{L}_{\text{KD},s}^{\text{freq}}=
\left\|S_{s,\text{low}}-T_{s,\text{low}}\right\|_1+
\alpha\left\|S_{s,\text{high}}-T_{s,\text{high}}\right\|_1,
\end{equation}
}
\rev{where $\alpha$ controls the relative emphasis placed on high-frequency supervision.}

\rev{To concentrate this supervision on spatial locations that are most difficult for the compact student, we construct a focal weighting map from the full teacher--student output disagreement at each scale. For pixel location $(i,j)$, the disagreement is computed as the mean absolute RGB difference}
{\ifshowrevisions\color{revblue}\fi
\begin{equation}
d_s(i,j)=\frac{1}{C}\sum_{c=1}^{C}\left|T_s(c,i,j)-S_s(c,i,j)\right|,
\end{equation}
}
\rev{where $C=3$ for RGB output. The disagreement map is min--max normalized independently for each sample and converted into the focal weight}
{\ifshowrevisions\color{revblue}\fi
\begin{equation}
w_s(i,j)=\left(\frac{d_s(i,j)-d_s^{\min}}{d_s^{\max}-d_s^{\min}+\epsilon}\right)^{\gamma},
\end{equation}
}
\rev{where $\gamma$ controls the concentration of the weighting function and $\epsilon$ prevents numerical instability. The focal map is detached from the computational graph before it is applied to the high-frequency loss. This stop-gradient operation is important because $w_s$ is itself computed from the teacher--student disagreement and therefore depends on the student prediction. If gradients were allowed to propagate through $w_s$, optimization would act both on the reconstruction error and on the mechanism that determines its spatial weight, introducing an additional pathway through which the weighted objective could change by modifying the weighting map itself. By detaching $w_s$, the disagreement observed at the current optimization step is treated as a fixed importance map, and gradients are directed only toward reducing the underlying student reconstruction mismatch. The map therefore determines \emph{where} greater supervision is applied without itself becoming an optimization target. Importantly, the weighting map is derived from the full teacher--student RGB disagreement, while it is applied specifically to the high-frequency residual mismatch. The resulting scale-specific frequency--focal objective is}
{\ifshowrevisions\color{revblue}\fi
\begin{equation}
\mathcal{L}_{\text{KD},s}^{\text{FF}}=
\left\|S_{s,\text{low}}-T_{s,\text{low}}\right\|_1+
\alpha\,\operatorname{mean}\!\left[
w_s\odot\left|S_{s,\text{high}}-T_{s,\text{high}}\right|
\right].
\end{equation}
}

\rev{The final frequency--focal distillation loss is obtained by averaging the scale-specific objectives:}
{\ifshowrevisions\color{revblue}\fi
\begin{equation}
\mathcal{L}_{\text{KD}}^{\text{FF}}=
\frac{1}{|\mathcal{S}|}
\sum_{s\in\mathcal{S}}
\mathcal{L}_{\text{KD},s}^{\text{FF}},
\qquad
\mathcal{S}=\{1,2,4\}.
\end{equation}
}

\rev{All frequency--focal experiments use a $5\times5$ Gaussian kernel with $\sigma=1.0$, $\alpha=3$, $\gamma=1$, and a maximum $\lambda_{\text{KD}}=0.3$ reached through a 10-epoch linear ramp. These settings are fixed for both tiny variants and all sequences; Section~\ref{sec:experiments} evaluates sensitivity to $\alpha$, kernel size, and $\sigma$. Conceptually, the Gaussian decomposition determines \emph{what} is transferred (structure versus detail), while the detached focal map determines \emph{where} the high-frequency term is emphasized, so difficult regions receive more supervision without changing inference complexity.}

\rev{For comparison, we also evaluate temporal-derivative and feature-level distillation. Temporal-derivative KD aligns frame-to-frame differences,}
{\ifshowrevisions\color{revblue}\fi
\begin{equation}
\mathcal{L}_{\text{KD}}^{\text{temp}} =
\| (S(\tau) - S(\tau-\Delta)) - (T(\tau) - T(\tau-\Delta)) \|_1 ,
\end{equation}}
\rev{while feature-level KD aligns intermediate activations using lightweight $1\times1$ channel adapters. These alternatives probe complementary supervision signals but are more sensitive to temporal or architectural mismatch than direct output-based transfer. The experiments in Section~\ref{sec:experiments} compare these strategies on BBB and validate the strongest frequency--focal formulation across the complete UVG benchmark. We next study low-precision inference as an orthogonal deployment constraint.}

\subsection{Low-Precision Inference and Quantization-Aware Training}

\rev{Numerical precision provides an efficiency axis complementary to width scaling because the learned weights constitute the stored video representation. We therefore use weights-only quantization: convolutional and linear weights are quantized per output channel, while activations, nonlinearities, and reconstructed RGB outputs remain in FP32. For a full-precision weight tensor $w$, each output channel is mapped using uniform min--max quantization}
{\ifshowrevisions\color{revblue}\fi
\begin{equation}
\tilde{w}
=
w_{\min}
+
s\,\operatorname{round}\!\left(\frac{w-w_{\min}}{s}\right),
\qquad
s=\frac{w_{\max}-w_{\min}}{2^b},
\end{equation}}
\rev{where $b$ is the target bit-width and $w_{\min}$, $w_{\max}$, and $s$ are computed independently per output channel.}

\rev{For post-training quantization (PTQ), this mapping is applied after optimization without updating model parameters. The experiments simulate reduced-precision weight storage rather than native integer execution: the quantized values are dequantized to their discrete levels and used with FP32 activations,}
{\ifshowrevisions\color{revblue}\fi
\begin{equation}
y=\mathcal{F}\!\left(x_{\mathrm{FP32}},\tilde{w}\right).
\end{equation}}
\rev{Thus, the reported INT8, INT6, and INT4 results isolate reconstruction sensitivity to weight precision and stored-weight compression; they are not measurements of native low-bit throughput or energy consumption.}

\rev{We evaluate QAT at INT8, INT6, and INT4. QAT fine-tunes the corresponding FP32 checkpoint while exposing the model to the same weight discretization during each forward pass, and a straight-through estimator keeps the underlying FP32 weights trainable:}
{\ifshowrevisions\color{revblue}\fi
\begin{equation}
w_{\mathrm{fq}}
=
w+\operatorname{sg}\!\left(\tilde{w}-w\right),
\end{equation}}
\rev{where $\operatorname{sg}(\cdot)$ denotes stop-gradient. Activations remain unquantized, and QAT changes only the training dynamics rather than the network structure.}

\rev{At INT4 we compare QAT-only against KD+QAT under otherwise identical conditions. QAT-only uses the original ground-truth reconstruction objective, whereas KD+QAT additionally uses a frozen FP32 NeRV-L teacher and final-output MSE supervision:}
{\ifshowrevisions\color{revblue}\fi
\begin{equation}
\mathcal{L}_{\mathrm{KD+QAT}}
=
\mathcal{L}_{\mathrm{GT}}
+
\lambda_{\mathrm{QKD}}
\left\|
S_{\mathrm{INT4}}(\tau)-T_{\mathrm{FP32}}(\tau)
\right\|_2^2,
\end{equation}}
\rev{with $\lambda_{\mathrm{QKD}}=1.0$. Keeping initialization, fake quantization, reconstruction loss, and optimization fixed isolates whether teacher supervision remains beneficial once the student is already adapting to severe numerical discretization.}

\section{Experimental Evaluation}
\label{sec:experiments}

\subsection{Datasets and Evaluation Protocol}

\begin{figure}[!t]
    \centering
    \includegraphics[width=0.75\linewidth,height=0.45\textheight,keepaspectratio]{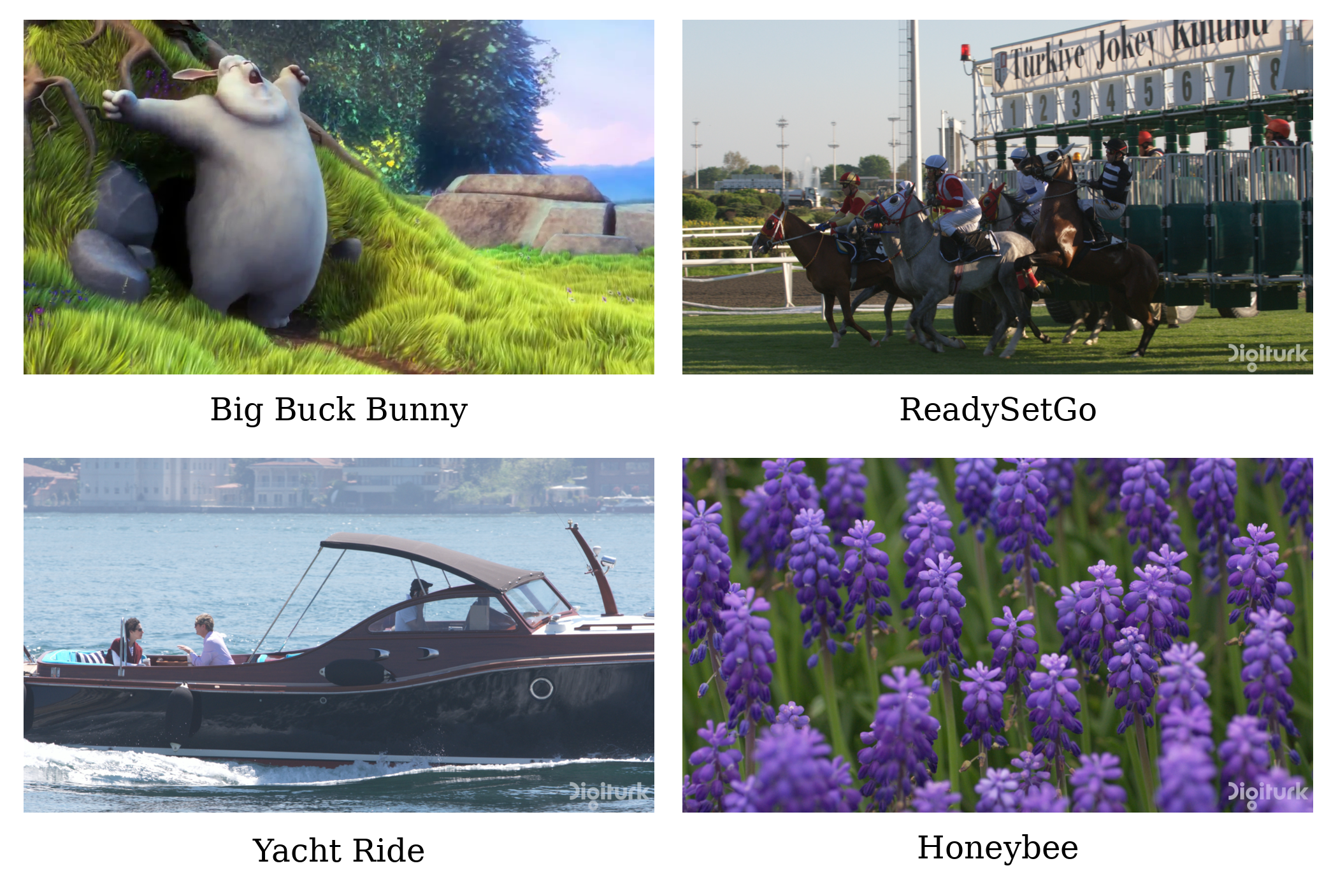}
    \caption{\rev{Representative frames from four evaluated videos. The complete study comprises BBB and all seven UVG sequences: Beauty, Bosphorus, HoneyBee, Jockey, ReadySetGo, ShakeNDry, and YachtRide.}}
    \label{fig:dataset_frames}
\end{figure}

\rev{We evaluate the proposed models on eight videos: Big Buck Bunny (BBB), which is used extensively in the foundational NeRV study~\cite{chen2021nerv}, and the complete seven-sequence UVG benchmark---Beauty, Bosphorus, HoneyBee, Jockey, ReadySetGo, ShakeNDry, and YachtRide~\cite{mercat2020uvg}. For controlled comparison, every evaluated sequence contains 132 frames and is reconstructed at the same spatial resolution of $720\times1280$. A separate NeRV model is trained for each video, but the architecture definition of each model variant, frame sampling, preprocessing, optimization protocol, distillation settings, and quantization procedure are held fixed across sequences. No sequence-specific hyperparameter tuning is performed. Representative frames from a subset of the evaluated videos are shown in Fig.~\ref{fig:dataset_frames}.}

\rev{For every sequence, baseline models are trained for 300 epochs with batch size 1 using Adam optimization, an initial learning rate of $5\times10^{-4}$, cosine learning-rate decay with a 20\% warmup period, Swish activations, and the Fusion6 reconstruction loss. Frames are sampled densely with a frame gap of 1 for both training and evaluation. These settings are identical for all eight sequences and for a given model variant; only the input video content changes. When extended training is evaluated, the duration is increased to 600 epochs while the remaining training configuration is preserved. This controlled protocol ensures that cross-sequence differences in reconstruction quality reflect video content rather than differences in frame count, resolution, optimization, or sequence-specific tuning. Reconstruction quality is measured using Peak Signal-to-Noise Ratio (PSNR) and Multi-Scale Structural Similarity (MS-SSIM) at the final reconstruction stage. Computational efficiency is reported using parameter count and analytically derived convolutional GFLOPs, while decoding throughput is measured in frames per second (FPS) using batch-size-one inference on an NVIDIA L4 GPU.}

\rev{To complement the compute-oriented metrics with a rate--distortion view, we also report an ideal packed model rate in bits per reconstructed pixel (bpp). For a model containing $N_{\mathrm{params}}$ parameters represented using $b$ bits per parameter over $N_f$ reconstructed frames of spatial resolution $H\times W$, the nominal model rate is defined as}
{\ifshowrevisions\color{revblue}\fi
\begin{equation}
R_{\mathrm{model}}(b)
=
\frac{N_{\mathrm{params}}\,b}
{N_f H W}
\quad \text{bits/pixel}.
\end{equation}}
\rev{All evaluated sequences use $N_f=132$ frames at $720\times1280$, so the same model configuration has an identical nominal rate across videos. This metric follows the same ideal packed-parameter convention used for the stored-weight footprint analysis and excludes metadata, container overhead, and entropy coding; it is therefore interpreted as a model-storage rate rather than a conventional codec bitstream rate.}

\rev{For training-time analysis, we compare baseline training, extended training, and knowledge distillation using a fixed NeRV-L teacher. For deployment analysis, we evaluate weights-only PTQ and QAT at INT8, INT6, and INT4 precision. Quantization uses uniform min--max mapping with per-output-channel quantization for convolutional and linear weight tensors; activations remain in full precision. Unless otherwise stated, frequency--focal distillation uses $\alpha=3$, a $5\times5$ Gaussian kernel with $\sigma=1.0$, focal exponent $\gamma=1$, and output scales $\{1,2,4\}$; these settings are shared by NeRV-T and NeRV-T+ and remain fixed across all sequences.}

\rev{To isolate each design choice, experiments vary one factor at a time---model capacity, distillation strategy, or numerical precision---while holding the remaining conditions fixed. BBB serves as the controlled reference for detailed ablations and deployment analysis, and the complete UVG benchmark evaluates whether the resulting trends persist across diverse content.}

\subsection{Capacity Scaling on Big Buck Bunny}

We first analyze the behavior of tiny NeRV variants on the Big Buck Bunny (BBB) sequence, which serves as a controlled reference case for understanding efficiency--quality trade-offs in the low-capacity regime. Table~\ref{tab:bbb_scaling} reports full-precision results for NeRV-T, NeRV-T+, and the reference NeRV-S/M/L configurations.

\begin{table*}[!t]
\centering
\scriptsize
\caption{Capacity scaling of NeRV models on BBB under full-precision inference. FPS is measured using single-GPU inference. FP32 rate denotes the ideal packed model payload normalized by the 132 reconstructed $720\times1280$ frames, while PSNR/GFLOP measures reconstruction quality normalized by computational cost.}
\label{tab:bbb_scaling}
\resizebox{\textwidth}{!}{%
\begin{tabular}{lccccccc}
\hline
Model & Params (M) & FLOPs (G) & FPS & PSNR (dB) & MS-SSIM & FP32 Rate (bpp) & PSNR/GFLOP \\
\hline
NeRV-T   & 0.80  & 22.62  & 197.9 & 27.84 & 0.882 & 0.210 & \textbf{1.231} \\
NeRV-T+  & 1.68  & 89.80  & 82.7  & 29.35 & 0.913 & 0.442 & 0.327 \\
NeRV-S~\cite{chen2021nerv} & 3.20  & 201.9 & 47.5 & 32.11 & 0.959 & 0.842 & 0.159 \\
NeRV-M~\cite{chen2021nerv} & 6.28  & 202.9 & 46.6 & 36.02 & 0.984 & 1.652 & 0.178 \\
NeRV-L~\cite{chen2021nerv} & 12.53 & 204.2 & 45.3 & 39.70 & 0.993 & 3.296 & 0.194 \\
\hline
\end{tabular}
}
\end{table*}

\begin{figure}[!t]
    \centering
    \includegraphics[width=0.82\linewidth,height=6cm,keepaspectratio]{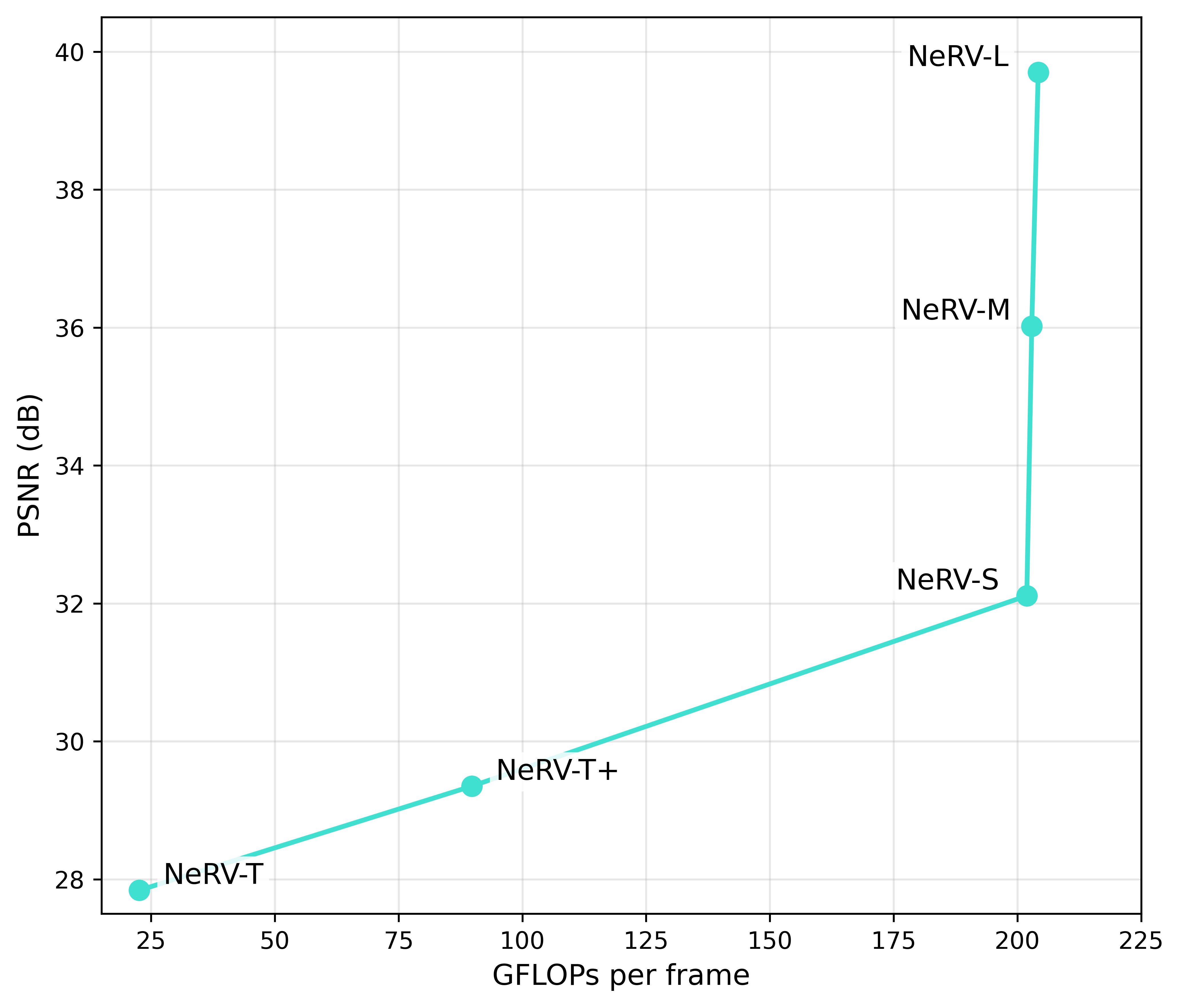}
    \caption{Quality--compute trade-off on BBB for NeRV model variants under full precision.}
    \label{fig:bbb_psnr_gflops}
\end{figure}

Figure~\ref{fig:bbb_psnr_gflops} visualizes the same scaling behavior as a quality--compute trade-off curve, highlighting the separation between the tiny operating points (NeRV-T/T+) and the high-compute NeRV-S/M/L regime.

\rev{NeRV-T defines the extreme-throughput point: 0.80M parameters and 22.62 GFLOPs yield 197.9 FPS, with the expected loss in PSNR and MS-SSIM from severe width reduction. NeRV-T+ increases high-resolution channel capacity, raising cost to 89.80 GFLOPs and reducing throughput to 82.7 FPS, but gains about 1.5 dB over NeRV-T. It remains more than twice as fast as NeRV-S and uses less than half its FLOPs, illustrating a useful intermediate quality--compute operating point.}

\rev{NeRV-S/M/L achieve higher absolute fidelity but require roughly 200 GFLOPs and decode below 50 FPS on the same hardware. The tiny variants therefore trade absolute quality for substantially lower computation; NeRV-T also has the highest reported PSNR/GFLOP in Table~\ref{tab:bbb_scaling}. Together, these results show that controlled width allocation provides a direct way to navigate the low-capacity efficiency--quality frontier.}

\rev{The same capacity scaling also defines a clear model rate--distortion progression. At FP32, NeRV-T operates at an ideal packed rate of 0.210~bpp, increasing to 0.442~bpp for NeRV-T+, 0.842~bpp for NeRV-S, 1.652~bpp for NeRV-M, and 3.296~bpp for NeRV-L. Moving from NeRV-T to NeRV-T+ approximately doubles the model rate but improves BBB PSNR by 1.51~dB, while NeRV-T+ still operates at only about half the rate of NeRV-S. The tiny variants therefore occupy distinct low-rate operating points in addition to their substantially lower computational cost.}

Table~\ref{tab:complexity_comparison} compares the computational complexity and parameter count of the proposed tiny NeRV variants with representative neural video representation methods from the literature. Existing approaches typically operate with approximately three million parameters and require around 180 to 360 GFLOPs per frame during decoding. In contrast, NeRV-T and NeRV-T+ substantially reduce both model size and computational cost, highlighting the suitability of the tiny NeRV regime for constrained deployment scenarios.

\begin{table}[!t]
\centering
\small
\caption{Comparison of computational complexity and parameter count with representative neural video representation methods reported in the literature.}
\label{tab:complexity_comparison}
\begin{tabular}{lcc}
\hline
Method & GFLOPs & Parameters \\
\hline
NeRV (Baseline)~\cite{chen2021nerv} & 201.9 & 3.20M \\
E-NeRV~\cite{kim2022enerv} & 208.0 & 3.31M \\
D-NeRV~\cite{zhao2023dnervmodelinginherentdynamics} & 181.0 & 3.00M \\
HiNeRV~\cite{hinerv2023} & 190.0 & 3.25M \\
FFNeRV~\cite{ffnerv2023} & 204.0 & 3.19M \\
SNeRV-B~\cite{kim2025snerv} & 363.6 & 3.00M \\
SNeRV-T~\cite{kim2025snerv} & 206.6 & 3.00M \\
HNeRV~\cite{li2023hnerv} & 188.0 & 3.28M \\
\textbf{NeRV-T (proposed)} & \textbf{22.62} & \textbf{0.80M} \\
\textbf{NeRV-T+ (proposed)} & \textbf{89.80} & \textbf{1.68M} \\
\hline
\end{tabular}
\end{table}

\rev{To translate these architectural reductions into concrete deployment requirements, we additionally express parameter count as stored-weight memory. For a model containing $N_{\mathrm{params}}$ parameters represented using $b$ bits per weight, the ideal packed-weight footprint is $N_{\mathrm{params}}b/8$ bytes. NeRV-T therefore requires approximately 3.20~MB, 0.80~MB, and 0.40~MB at FP32, INT8, and INT4 precision, respectively, while NeRV-T+ requires 6.72~MB, 1.68~MB, and 0.84~MB. For comparison, NeRV-S requires 12.80~MB, 3.20~MB, and 1.60~MB at the same precisions. These reductions are substantial enough to cross practical memory-capacity thresholds on representative embedded hardware, while the accompanying reduction in per-frame GFLOPs addresses the compute constraints of accelerator-equipped edge platforms.}

\begin{table*}[!t]
\centering
\footnotesize
\renewcommand{\arraystretch}{1.08}
\caption{\rev{Representative embedded and edge-AI hardware constraints in relation to TinyNeRV stored-weight footprints and computational requirements. Hardware specifications are obtained from manufacturer datasheets and official technical specifications~\cite{stm32h753_datasheet,stm32n6_datasheet,jetson_orin_nano_datasheet,jetson_orin_nano_super_specs}.}}
\label{tab:deployment_platforms}
\begin{tabular}{p{2.6cm}p{2.3cm}p{3.0cm}p{2.8cm}p{5.0cm}}
\hline
\rev{Platform} & \rev{Class} & \rev{Memory} & \rev{Compute / Power} & \rev{Relation to TinyNeRV} \\
\hline
\rev{STM32H753XI} & \rev{Cortex-M7 MCU} & \rev{1~MB RAM; 2~MB Flash} & \rev{480~MHz CPU} & \rev{NeRV-T at INT8 (0.80~MB) and INT4 (0.40~MB), and NeRV-T+ at INT4 (0.84~MB), have stored-weight footprints below the 1~MB RAM capacity.} \\
\rev{STM32N657} & \rev{AI-enabled MCU} & \rev{4.2~MB SRAM} & \rev{Neural-ART accelerator; up to 600~GOPS} & \rev{NeRV-T at FP32 (3.20~MB), NeRV-T and NeRV-T+ at INT8 (0.80/1.68~MB), and both variants at INT4 (0.40/0.84~MB) have stored-weight footprints below the available SRAM capacity.} \\
\rev{Jetson Orin Nano Super} & \rev{Edge-AI GPU} & \rev{8~GB LPDDR5; 102~GB/s} & \rev{67 INT8 TOPS; 7--25~W} & \rev{All evaluated NeRV weight footprints are well within device memory; the reduction from 201.9~GFLOPs/frame for NeRV-S to 89.80 and 22.62~GFLOPs/frame for NeRV-T+ and NeRV-T reduces the computational demand of real-time edge inference.} \\
\hline
\end{tabular}
\end{table*}

\rev{Table~\ref{tab:deployment_platforms} shows that TinyNeRV crosses meaningful hardware thresholds rather than providing only relative reductions. At INT8, NeRV-T reduces the 3.20~MB NeRV-S weight footprint to 0.80~MB, placing it within the memory range of 1~MB-class MCUs; the STM32N657's 4.2~MB SRAM can accommodate NeRV-T at FP32 and both tiny variants at INT8 or INT4. On the Jetson Orin Nano Super, memory is less restrictive, but compute remains relevant: relative to NeRV-S, NeRV-T+ and NeRV-T reduce per-frame GFLOPs by approximately 55.5\% and 88.8\%, respectively. These comparisons motivate NeRV-T for the tightest budgets and NeRV-T+ when additional computation can be exchanged for higher fidelity.}

\rev{Overall, NeRV-T prioritizes minimum compute and maximum throughput, while NeRV-T+ provides a higher-fidelity point at moderate additional cost. The following analyses examine whether training-time supervision and reduced precision can further improve these operating points.}

\subsection{Temporal Stability Analysis}

\rev{In addition to spatial reconstruction quality, we evaluate temporal consistency using temporal PSNR (T-PSNR) and temporal SSIM (tSSIM). For each consecutive frame pair, temporal difference images $(I_t-I_{t-1})$ are formed for the ground-truth and reconstructed sequences, and PSNR/SSIM are computed between the corresponding difference signals. Higher values therefore indicate that the reconstructed sequence more accurately preserves the frame-to-frame changes present in the ground truth.}

\rev{Table~\ref{tab:temporal_stability} reports this analysis for all eight evaluated videos. A consistent capacity ordering is observed on every sequence: NeRV-T $<$ NeRV-T+ $<$ NeRV-S $<$ NeRV-M $<$ NeRV-L for both temporal metrics. NeRV-T+ improves over NeRV-T on all eight videos, with an average gain of 0.477~dB in T-PSNR and 0.0088 in tSSIM. The largest T-PSNR gains occur on Jockey (+1.588~dB) and ReadySetGo (+1.079~dB), whereas the two tiny variants are much closer on HoneyBee (+0.022~dB) and ShakeNDry (+0.110~dB).}

\begin{table*}[!t]
\centering
\tiny
\renewcommand{\arraystretch}{1.05}
\caption{\rev{Temporal stability across all eight evaluated videos under FP32 inference.}}
\label{tab:temporal_stability}
{\ifshowrevisions\color{revblue}\fi
\resizebox{\textwidth}{!}{%
\begin{tabular}{lcccccccccc}
\hline
& \multicolumn{2}{c}{NeRV-T} & \multicolumn{2}{c}{NeRV-T+} & \multicolumn{2}{c}{NeRV-S} & \multicolumn{2}{c}{NeRV-M} & \multicolumn{2}{c}{NeRV-L} \\
\cline{2-11}
Sequence & T-PSNR & tSSIM & T-PSNR & tSSIM & T-PSNR & tSSIM & T-PSNR & tSSIM & T-PSNR & tSSIM \\
\hline
BBB        & 33.115 & 0.9390 & 33.564 & 0.9426 & 37.369 & 0.9701 & 37.640 & 0.9739 & 40.726 & 0.9834 \\
HoneyBee   & 38.334 & 0.9672 & 38.356 & 0.9673 & 38.390 & 0.9674 & 38.613 & 0.9678 & 38.994 & 0.9687 \\
ReadySetGo & 23.171 & 0.7900 & 24.250 & 0.8224 & 26.915 & 0.8838 & 29.717 & 0.9207 & 32.069 & 0.9448 \\
YachtRide  & 26.431 & 0.8343 & 26.624 & 0.8400 & 27.229 & 0.8563 & 28.849 & 0.8901 & 31.259 & 0.9309 \\
Beauty     & 30.452 & 0.8548 & 30.667 & 0.8576 & 31.324 & 0.8662 & 32.054 & 0.8785 & 32.559 & 0.8906 \\
Bosphorus  & 33.471 & 0.9238 & 33.632 & 0.9260 & 34.197 & 0.9327 & 35.479 & 0.9467 & 36.826 & 0.9596 \\
Jockey     & 24.683 & 0.8492 & 26.271 & 0.8718 & 29.510 & 0.9070 & 32.304 & 0.9320 & 34.681 & 0.9536 \\
ShakeNDry  & 29.345 & 0.8404 & 29.455 & 0.8416 & 29.655 & 0.8451 & 30.496 & 0.8713 & 32.732 & 0.9162 \\
\hline
Average    & 29.875 & 0.8748 & 30.352 & 0.8837 & 31.824 & 0.9036 & 33.144 & 0.9226 & 34.981 & 0.9435 \\
\hline
\end{tabular}%
}}
\end{table*}

\rev{Because NeRV-T and NeRV-T+ use the same temporal positional encoding and contain no explicit recurrent or temporal-consistency module, the improved temporal metrics of NeRV-T+ are attributable to its greater representational capacity rather than a dedicated temporal mechanism. The temporal results therefore reinforce the spatial scaling analysis: increasing capacity within the tiny regime improves not only per-frame reconstruction quality but also the accuracy with which neighboring temporal coordinates are mapped to their corresponding frame changes.}

\subsection{Training-Time Improvements on Big Buck Bunny}

We next examine whether training-time strategies can improve the reconstruction quality of tiny NeRV models without increasing inference complexity. In addition to baseline training for 300 epochs, we evaluate extended training (600 epochs) and four knowledge distillation variants using a fixed NeRV-L teacher: final-output distillation, frequency--focal distillation, temporal-difference distillation, and feature-level distillation. All results are reported under full-precision inference.

\begin{table}[!t]
\centering
\small
\renewcommand{\arraystretch}{1.1}
\caption{Training-time improvements for NeRV-T on BBB under full-precision inference. PSNR gain is measured relative to the 300-epoch baseline.}
\label{tab:bbb_kd_T}
\begin{tabular}{p{3.0cm}ccc}
\hline
Method & PSNR & MS-SSIM & $\Delta\mathrm{PSNR}$ \\
\hline
Baseline (300 ep.)        & 27.84 & 0.882 & -- \\
Extended Training (600 ep.) & 28.19 & 0.891 & +0.35 \\
KD Feature-Level          & 27.98 & 0.884 & +0.14 \\
KD Temporal Difference    & 28.51 & 0.896 & +0.67 \\
KD Final-Output           & 28.60 & 0.897 & +0.76 \\
\textbf{KD Frequency--Focal} & \textbf{28.68} & \textbf{0.898} & \textbf{+0.84} \\
\hline
\end{tabular}
\end{table}

\begin{table}[!t]
\centering
\small
\renewcommand{\arraystretch}{1.1}
\caption{Training-time improvements for NeRV-T+ on BBB under full-precision inference. PSNR gain is measured relative to the 300-epoch baseline.}
\label{tab:bbb_kd_Tplus}
\begin{tabular}{p{3.1cm}ccc}
\hline
Method & PSNR & MS-SSIM & $\Delta\mathrm{PSNR}$ \\
\hline
Baseline (300 epochs) &
29.35 & 0.913 & -- \\

Extended Training (600 epochs) &
29.64 & 0.916 & +0.29 \\

KD Feature-Level &
29.45 & 0.909 & +0.10 \\

KD Temporal Difference &
29.82 & 0.918 & +0.47 \\

KD Final-Output &
30.05 & 0.922 & +0.70 \\

\textbf{KD Frequency--Focal} &
\textbf{30.23} & \textbf{0.923} & \textbf{+0.88} \\
\hline
\end{tabular}
\end{table}

\rev{Extended training yields only +0.35 dB for NeRV-T and +0.29 dB for NeRV-T+, whereas output-based distillation provides substantially larger gains. Final-output KD reaches +0.76/+0.70 dB, and frequency--focal KD performs best at +0.84/+0.88 dB for NeRV-T/T+, respectively. Temporal-difference KD also improves both students, while feature-level KD provides only marginal gains, suggesting that direct output supervision is better matched to severely width-constrained decoders than intermediate feature alignment.}

\rev{To isolate the two components, we compare final-output KD with frequency-only, focal-only, and combined frequency--focal variants. Frequency-only KD keeps the low/high-frequency decomposition without focal weighting, whereas focal-only KD applies disagreement weighting without frequency decomposition.}

\begin{table}[!h]
\centering
\footnotesize
\renewcommand{\arraystretch}{1.08}
\caption{\rev{Component ablation of frequency--focal knowledge distillation on BBB.}}
\label{tab:ff_component_ablation}
{\ifshowrevisions\color{revblue}\fi
\begin{tabular}{lcccc}
\hline
Method & Freq. & Focal & NeRV-T & NeRV-T+ \\
\hline
Final-output KD       & --  & --  & 28.60 & 30.05 \\
Frequency-only KD     & \checkmark & -- & 28.64 & 30.13 \\
Focal-only KD         & -- & \checkmark & 28.62 & 30.10 \\
Frequency--Focal KD   & \checkmark & \checkmark & \textbf{28.68} & \textbf{30.23} \\
\hline
\end{tabular}}
\end{table}

\rev{Both components provide independent gains over final-output KD, and their combination is strongest for both students. Frequency decomposition determines which reconstruction component receives detail-oriented supervision, while the focal map determines where that supervision is concentrated, confirming that the two mechanisms are complementary.}

\rev{Table~\ref{tab:ff_sensitivity} evaluates sensitivity to $\alpha$ and the Gaussian decomposition parameters on BBB. With $k=5$ and $\sigma=1.0$, performance improves as $\alpha$ increases from 1 to 3 and then declines slightly, making $\alpha=3$ best for both students. The response is smooth: moving from $\alpha=3$ to 2 or 4 changes PSNR by at most 0.04~dB, indicating limited sensitivity around the optimum.}

\begin{table}[!h]
\centering
\footnotesize
\renewcommand{\arraystretch}{1.05}
\caption{\rev{Sensitivity of frequency--focal distillation to the high-frequency weight $\alpha$ and Gaussian decomposition parameters on BBB. The default configuration is $k=5$, $\sigma=1.0$, and $\alpha=3$.}}
\label{tab:ff_sensitivity}
\begin{tabular}{cccccc}
\hline
\rev{$k$} & \rev{$\sigma$} & \rev{$\alpha$} & \rev{Setting} & \rev{NeRV-T} & \rev{NeRV-T+} \\
 & & & & \rev{PSNR} & \rev{PSNR} \\
\hline
\rev{5} & \rev{1.0} & \rev{1} & \rev{Low $\alpha$} & \rev{28.57} & \rev{30.11} \\
\rev{5} & \rev{1.0} & \rev{2} & \rev{Moderate $\alpha$} & \rev{28.64} & \rev{30.19} \\
\rev{5} & \rev{1.0} & \rev{\textbf{3}} & \rev{\textbf{Default}} & \rev{\textbf{28.68}} & \rev{\textbf{30.23}} \\
\rev{5} & \rev{1.0} & \rev{4} & \rev{Higher $\alpha$} & \rev{28.66} & \rev{30.21} \\
\rev{5} & \rev{1.0} & \rev{5} & \rev{Strong $\alpha$} & \rev{28.62} & \rev{30.17} \\
\hline
\rev{3} & \rev{1.0} & \rev{3} & \rev{Smaller kernel} & \rev{28.60} & \rev{30.15} \\
\rev{7} & \rev{1.0} & \rev{3} & \rev{Larger kernel} & \rev{28.65} & \rev{30.21} \\
\rev{5} & \rev{0.5} & \rev{3} & \rev{Narrow Gaussian} & \rev{28.58} & \rev{30.13} \\
\rev{5} & \rev{1.5} & \rev{3} & \rev{Wide Gaussian} & \rev{28.63} & \rev{30.18} \\
\hline
\end{tabular}
\end{table}

\rev{The Gaussian parameters are similarly stable: changing the kernel from 5 to 3 or 7 alters PSNR by at most 0.08~dB, and varying $\sigma$ from 1.0 to 0.5 or 1.5 produces only modest changes. The default $k=5$, $\sigma=1.0$, $\alpha=3$ remains best for both capacities and is therefore used throughout the remaining experiments without sequence-specific tuning.}

\rev{To place the proposed distilled tiny models in context, Table~\ref{tab:efficient_nerv_comparison} compares representative efficiency-oriented NeRV variants. Since frequency--focal distillation improves reconstruction quality without modifying the student architecture or inference-time computation, the distilled NeRV-T and NeRV-T+ models are used as the strongest full-precision TinyNeRV configurations. Binary-NeRV and Ghost-NeRV pursue efficiency through complementary mechanisms: hybrid numerical binarization and operator-level convolutional redesign, respectively, whereas TinyNeRV explores a lower-capacity regime through controlled width scaling followed by training-time quality recovery. Because the compared studies use different arithmetic-complexity conventions, Table~\ref{tab:efficient_nerv_comparison} distinguishes conventional GFLOPs from the FP32-equivalent binary complexity reported by Binary-NeRV. Hardware-dependent inference throughput is discussed separately below.}

\begin{table*}[!t]
\centering
\footnotesize
\renewcommand{\arraystretch}{1.05}
\caption{\rev{Comparison with recent efficiency-oriented NeRV variants. Conventional GFLOPs for NeRV, Ghost-NeRV, and TinyNeRV follow the standard $2\times$MAC counting convention. Binary-NeRV reports FP32-equivalent computational cost obtained from binary-operation conversion.}}
\label{tab:efficient_nerv_comparison}
\begin{tabular}{p{2.3cm}p{4.0cm}ccc}
\hline
Method & Strategy & Params (M) & GFLOPs & PSNR \\
\hline
NeRV~\cite{chen2021nerv}
& Full-precision decoder
& 3.20 & 201.9 & 32.11 \\
\hline
Binary-NeRV (L4)~\cite{shanableh2026binary}
& Hybrid-precision binarization
& 3.24 & 72.6$^\dagger$ & 30.41 \\
\hline
Ghost-NeRV (L3,L4)~\cite{shanableh2026ghost}
& Ghost convolutions
& 2.87 & 107.32 & 32.55 \\
\hline
NeRV-T + KD-FF (ours)
& Width scaling + KD-FF
& \textbf{0.80} & \textbf{22.62} & 28.68 \\
\hline
NeRV-T+ + KD-FF (ours)
& Width scaling + KD-FF
& 1.68 & 89.80 & 30.23 \\
\hline
\end{tabular}
\vspace{0.3em}
\begin{minipage}{0.98\textwidth}
\footnotesize
\rev{$^\dagger$ Binary-NeRV reports FP32-equivalent GFLOPs by converting binary operations using its stated BOP-to-FLOP equivalence; this value should therefore not be interpreted as identical to conventional floating-point GFLOPs.}
\end{minipage}
\end{table*}

\rev{The comparison shows that TinyNeRV occupies a distinct operating region. NeRV-T achieves the lowest parameter count and conventional computational cost among the compared configurations, while sacrificing reconstruction fidelity relative to larger or operator-optimized variants. NeRV-T+ provides a more balanced operating point: it remains substantially smaller than Ghost-NeRV and requires lower conventional computational cost, while achieving PSNR comparable to the selected Binary-NeRV operating point with considerably fewer parameters. These results position TinyNeRV as complementary to Binary-NeRV and Ghost-NeRV: rather than reducing arithmetic precision or replacing standard convolutions, TinyNeRV characterizes how far NeRV-style decoders can be pushed into the low-capacity regime while retaining useful reconstruction quality.}

\rev{\textit{Inference throughput.} We additionally report FPS as implementation-specific deployment context, while separating it from Table~\ref{tab:efficient_nerv_comparison} because the available measurements were obtained on different hardware platforms. Under our common batch-size-one NVIDIA L4 evaluation setup, NeRV-S, NeRV-T+, and NeRV-T achieve 47.5, 82.7, and 197.9~FPS, respectively. These measurements provide a controlled within-platform comparison and show that the reduced computational cost of the tiny variants translates into higher decoding throughput in the same implementation. Ghost-NeRV reports 51.22~FPS on an NVIDIA A10G GPU~\cite{shanableh2026ghost}; this value is retained only as platform-specific context and should not be compared numerically with the L4 measurements. Binary-NeRV does not provide a directly comparable native-binary throughput measurement. In particular, its available PyTorch implementation represents binary weights and activations but executes the convolutions using standard floating-point operators rather than packed-bit XNOR/popcount kernels, so benchmarking that implementation would primarily measure binary-emulation overhead rather than the throughput achievable with an optimized binary backend. Consequently, cross-method conclusions in Table~\ref{tab:efficient_nerv_comparison} are based primarily on parameter count, computational complexity, and reconstruction quality, whereas FPS is used only for controlled within-platform or study-specific deployment context.}

Overall, these results demonstrate that knowledge distillation is significantly more effective than extended training for improving tiny NeRV variants. Moreover, frequency-aware supervision consistently provides the strongest gains, reinforcing the importance of preserving high-frequency detail in the low-capacity regime. The next subsection investigates whether these gains persist under aggressive low-precision deployment.

\subsection{Low-Precision Deployment on Big Buck Bunny}

\rev{We next evaluate weights-only quantization on BBB. Because the video is stored in the model parameters, reducing weight precision provides a rate axis complementary to architectural scaling. Under the nominal packed-rate definition in Section~IV-A, NeRV-T decreases from 0.210~bpp at FP32 to 0.0526, 0.0395, and 0.0263~bpp at INT8, INT6, and INT4, respectively. The corresponding NeRV-T+ rates are 0.442, 0.1105, 0.0829, and 0.0552~bpp. We first analyze INT8 and INT6, followed by the more aggressive INT4 regime.}

\begin{table}[!t]
\centering
\caption{Moderate-precision deployment results on BBB using weights-only quantization. FP32 denotes the full-precision baseline without quantization.}
\label{tab:bbb_quant_mid}
\begin{tabular}{lcccc}
\hline
 & \multicolumn{2}{c}{NeRV-T (PSNR dB)} & \multicolumn{2}{c}{NeRV-T+ (PSNR dB)} \\
\cline{2-3} \cline{4-5}
Precision & PTQ & QAT & PTQ & QAT \\
\hline
FP32 (Baseline) & \multicolumn{2}{c}{27.84} & \multicolumn{2}{c}{29.35} \\
\hline
INT8 & 27.84 & 27.88 & 29.31 & 29.36 \\
INT6 & 26.92 & 27.63 & 28.78 & 29.10 \\
\hline
\end{tabular}
\end{table}

\rev{Table~\ref{tab:bbb_quant_mid} shows that INT8 PTQ is effectively lossless for NeRV-T and only 0.04~dB below FP32 for NeRV-T+, with little additional benefit from QAT. INT6 is more sensitive: PTQ loses 0.92~dB for NeRV-T and 0.57~dB for NeRV-T+, while QAT recovers most of the deficit. Thus, INT8 is a robust moderate-precision point, whereas INT6 already benefits from retraining-aware adaptation.}

\begin{table}[!t]
\centering
\small
\renewcommand{\arraystretch}{1.1}
\caption{INT4 quantization results on BBB (PSNR in dB). PTQ denotes post-training quantization, QAT denotes quantization-aware training, and KD+QAT combines knowledge distillation with quantization-aware training.}
\label{tab:bbb_quant_int4}
\begin{tabular}{lccc}
\hline
Model & PTQ & QAT & KD+QAT \\
\hline
NeRV-T  & 24.66 & 27.12 & \textbf{27.66} \\
NeRV-T+ & 26.31 & 28.72 & \textbf{29.25} \\
\hline
\end{tabular}
\end{table}

\rev{INT4 PTQ is much more damaging, reducing NeRV-T and NeRV-T+ by 3.18 and 3.04~dB relative to FP32. QAT recovers 2.46 and 2.41~dB, respectively. Adding the frozen teacher yields a further +0.54~dB for NeRV-T and +0.53~dB for NeRV-T+, showing that QAT and distillation provide complementary recovery under severe discretization. Overall, the BBB results identify a near-lossless INT8 regime, a transitional INT6 regime, and an INT4 regime in which retraining-aware adaptation is required to preserve reconstruction quality.}

\rev{Viewed from a rate--distortion perspective, quantization moves each fixed-capacity model horizontally toward substantially lower model rates, while QAT and KD+QAT recover reconstruction quality without increasing that deployed rate. For example, NeRV-T at INT4 operates at 0.0263~bpp whether evaluated using PTQ, QAT, or KD+QAT, but PSNR increases from 24.66 to 27.12 and 27.66~dB, respectively. Similarly, NeRV-T+ remains at 0.0552~bpp while improving from 26.31~dB under PTQ to 28.72~dB with QAT and 29.25~dB with KD+QAT. Thus, retraining shifts the distortion performance upward at a fixed low-bit model rate rather than increasing the representation size.}

\subsection{Cross-Sequence Validation on the Complete UVG Benchmark}

\rev{We next evaluate whether the capacity, distillation, and quantization trends observed on BBB persist across the complete seven-sequence UVG benchmark. All seven UVG sequences are evaluated over the same 132-frame span at $720\times1280$ resolution using the identical training and evaluation protocol defined in Section~IV-A; no architecture, optimization, distillation, or quantization hyperparameter is adjusted on a per-sequence basis. This controlled setup allows differences in reconstruction quality to be attributed to video content rather than to changes in frame count, resolution, or experimental configuration.}

\begin{figure*}[!t]
\centering
\includegraphics[width=\linewidth]{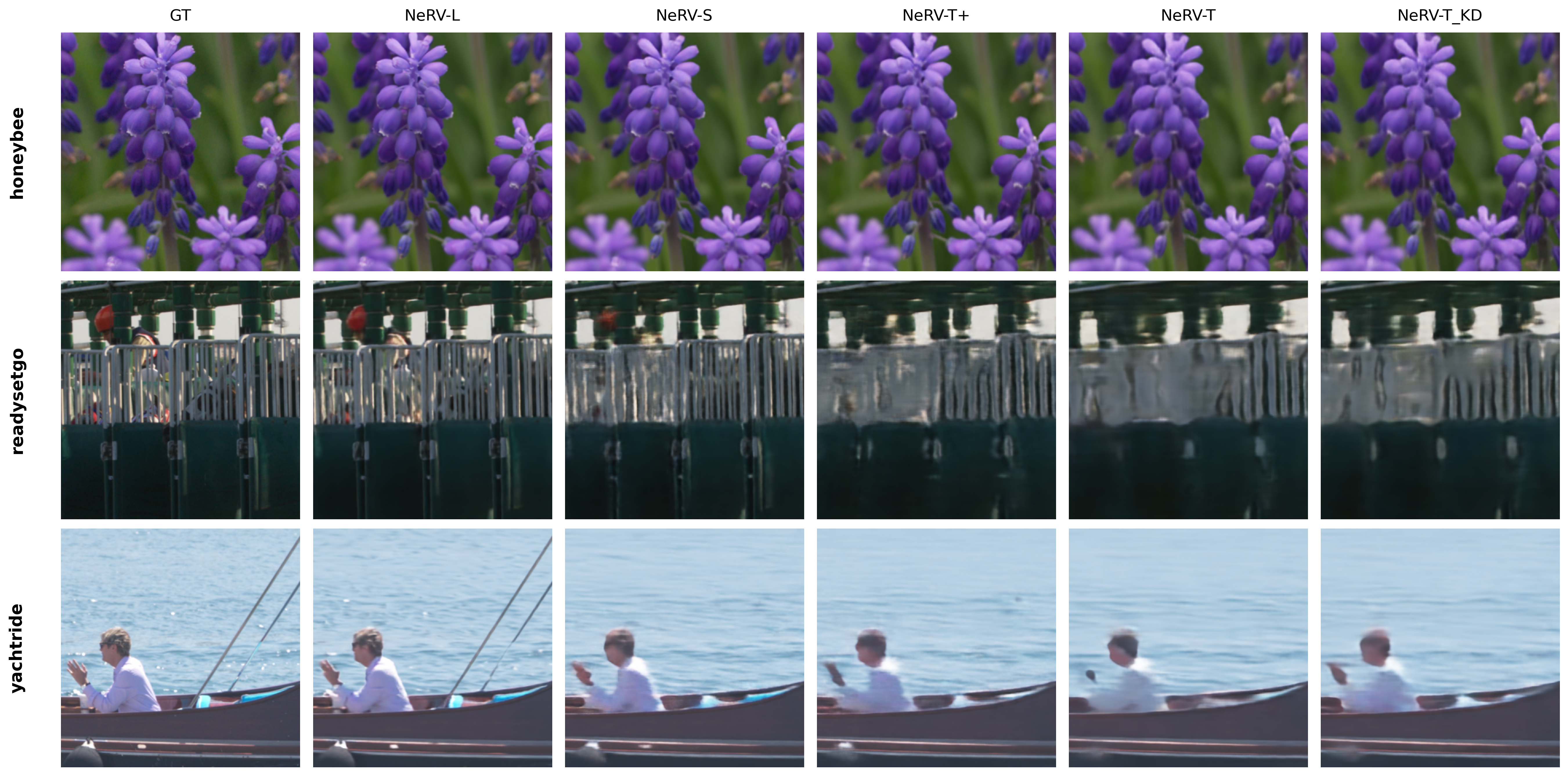}
\caption{\rev{Representative qualitative comparison on HoneyBee, ReadySetGo, and YachtRide. NeRV-L and NeRV-S provide higher-capacity references, while NeRV-T and NeRV-T+ illustrate aggressive capacity reduction and the distilled NeRV-T output shows the recovery obtained from knowledge distillation.}}
\label{fig:qualitative_crossseq}
\end{figure*}

\rev{Table~\ref{tab:cross_fp_kd} shows consistent full-precision trends across all seven UVG sequences. NeRV-T+ exceeds NeRV-T on every sequence by 1.36~dB on average. Frequency--focal KD also improves every sequence, averaging +0.70~dB for NeRV-T and +0.78~dB for NeRV-T+, and exceeds extended fine-tuning by +0.31/+0.33~dB on average. The largest KD gains occur on Jockey (+1.30/+1.52~dB), while average MS-SSIM rises from 0.9036 to 0.9128 for NeRV-T and from 0.9213 to 0.9305 for NeRV-T+.}

\begin{table*}[!t]
\centering
\tiny
\renewcommand{\arraystretch}{1.05}
\caption{\rev{Full-precision reconstruction across the complete UVG benchmark. Each entry reports PSNR (dB) / MS-SSIM. ``FT'' denotes extended fine-tuning and ``FF-KD'' denotes frequency--focal distillation.}}
\label{tab:cross_fp_kd}
{\ifshowrevisions\color{revblue}\fi
\resizebox{\textwidth}{!}{%
\begin{tabular}{lccccccccc}
\hline
& \multicolumn{3}{c}{Reference baselines} & \multicolumn{3}{c}{NeRV-T} & \multicolumn{3}{c}{NeRV-T+} \\
\cline{2-10}
Sequence & NeRV-S & NeRV-M & NeRV-L & Base & FT & FF-KD & Base & FT & FF-KD \\
\hline
HoneyBee   & 39.15/0.9914 & 40.75/0.9933 & 41.70/0.9944 & 34.26/0.9794 & 34.45/0.9789 & 34.79/0.9812 & 37.00/0.9879 & 37.22/0.9845 & 37.54/0.9887 \\
ReadySetGo & 28.45/0.9490 & 31.78/0.9754 & 34.42/0.9864 & 23.63/0.8608 & 23.83/0.8693 & 24.28/0.8759 & 25.07/0.8951 & 25.31/0.9022 & 25.75/0.9087 \\
YachtRide  & 29.03/0.9249 & 31.65/0.9595 & 34.23/0.9792 & 26.23/0.8785 & 26.34/0.8807 & 26.69/0.8855 & 27.23/0.8940 & 27.42/0.9008 & 27.77/0.9037 \\
Beauty     & 33.91/0.9319 & 34.84/0.9424 & 35.45/0.9517 & 32.24/0.9105 & 32.69/0.9166 & 32.98/0.9188 & 32.80/0.9182 & 33.27/0.9243 & 33.56/0.9263 \\
Bosphorus  & 33.65/0.9609 & 36.51/0.9815 & 38.95/0.9904 & 29.99/0.9103 & 30.44/0.9179 & 30.70/0.9182 & 31.21/0.9286 & 31.76/0.9363 & 31.97/0.9378 \\
Jockey     & 32.02/0.9437 & 35.12/0.9666 & 37.56/0.9814 & 26.96/0.8773 & 27.93/0.8937 & 28.26/0.8951 & 28.62/0.9049 & 29.82/0.9209 & 30.14/0.9223 \\
ShakeNDry  & 32.09/0.9311 & 33.20/0.9527 & 35.50/0.9763 & 30.33/0.9084 & 30.67/0.9131 & 30.85/0.9146 & 31.21/0.9202 & 31.49/0.9235 & 31.85/0.9259 \\
\hline
Average    & 32.61/0.9476 & 34.84/0.9673 & 36.83/0.9800 & 29.09/0.9036 & 29.48/0.9100 & 29.79/0.9128 & 30.45/0.9213 & 30.90/0.9275 & 31.23/0.9305 \\
\hline
\end{tabular}%
}}
\end{table*}

\rev{Table~\ref{tab:cross_quant_mid} shows the same moderate-precision behavior across UVG. INT8 PTQ is only 0.02~dB below FP32 on average for either tiny variant. INT6 PTQ loses 0.34~dB for NeRV-T and 0.33~dB for NeRV-T+, while QAT reduces the remaining deficits to 0.08 and 0.11~dB, respectively.}

\begin{table*}[!t]
\centering
\tiny
\renewcommand{\arraystretch}{1.05}
\caption{\rev{INT8 and INT6 weights-only quantization across the complete UVG benchmark (PSNR in dB).}}
\label{tab:cross_quant_mid}
{\ifshowrevisions\color{revblue}\fi
\resizebox{\textwidth}{!}{%
\begin{tabular}{lcccccccc}
\hline
& \multicolumn{4}{c}{NeRV-T} & \multicolumn{4}{c}{NeRV-T+} \\
\cline{2-9}
Sequence & INT8 PTQ & INT8 QAT & INT6 PTQ & INT6 QAT & INT8 PTQ & INT8 QAT & INT6 PTQ & INT6 QAT \\
\hline
HoneyBee   & 34.17 & 34.24 & 33.05 & 33.90 & 36.92 & 36.97 & 35.89 & 36.54 \\
ReadySetGo & 23.62 & 23.66 & 23.51 & 23.63 & 25.07 & 25.10 & 24.91 & 25.06 \\
YachtRide  & 26.23 & 26.35 & 26.11 & 26.23 & 27.22 & 27.24 & 27.09 & 27.20 \\
Beauty     & 32.24 & 32.28 & 32.04 & 32.20 & 32.79 & 32.81 & 32.55 & 32.72 \\
Bosphorus  & 29.99 & 30.00 & 29.82 & 29.95 & 31.19 & 31.22 & 30.98 & 31.11 \\
Jockey     & 26.95 & 27.02 & 26.89 & 26.99 & 28.61 & 28.69 & 28.53 & 28.63 \\
ShakeNDry  & 30.30 & 30.33 & 29.83 & 30.22 & 31.19 & 31.21 & 30.90 & 31.12 \\
\hline
Average    & 29.07 & 29.13 & 28.75 & 29.02 & 30.43 & 30.46 & 30.12 & 30.34 \\
\hline
\end{tabular}%
}}
\end{table*}

\rev{INT4 remains substantially harder. Relative to FP32, PTQ loses 2.99~dB for NeRV-T and 3.03~dB for NeRV-T+ on average; QAT recovers 2.05 and 1.92~dB, respectively. KD+QAT adds a further +0.22/+0.17~dB on average and improves over QAT in 13 of 14 sequence--model cases. The only exception is NeRV-T on Bosphorus ($-0.02$~dB), showing that teacher supervision is beneficial on average but content-dependent at INT4.}

\rev{The complete UVG results also reveal how the proposed training strategies affect model rate--distortion behavior. At full precision, frequency--focal distillation leaves the model rate unchanged while increasing average PSNR from 29.09 to 29.79~dB for NeRV-T at 0.210~bpp and from 30.45 to 31.23~dB for NeRV-T+ at 0.442~bpp. Precision reduction then moves these models to substantially lower rates: INT8 PTQ reaches 0.0526/0.1105~bpp for NeRV-T/T+ with only about 0.02~dB average loss relative to FP32, while INT4 reduces the rates further to 0.0263/0.0552~bpp. At these INT4 rates, KD+QAT achieves average PSNR values of 28.37 and 29.50~dB, respectively. Capacity and precision scaling therefore control movement along the model-rate axis, whereas distillation and QAT primarily improve distortion at a fixed rate.}

\begin{table*}[!t]
\centering
\footnotesize
\renewcommand{\arraystretch}{1.05}
\caption{\rev{INT4 weights-only deployment across the complete UVG benchmark (PSNR in dB).}}
\label{tab:cross_quant_int4}
{\ifshowrevisions\color{revblue}\fi
\begin{tabular}{lcccccc}
\hline
& \multicolumn{3}{c}{NeRV-T} & \multicolumn{3}{c}{NeRV-T+} \\
\cline{2-7}
Sequence & PTQ & QAT & KD+QAT & PTQ & QAT & KD+QAT \\
\hline
HoneyBee   & 26.54 & 31.19 & 31.70 & 29.22 & 33.67 & 33.91 \\
ReadySetGo & 22.13 & 23.29 & 23.52 & 23.26 & 24.61 & 24.88 \\
YachtRide  & 24.84 & 25.95 & 26.32 & 25.69 & 26.77 & 26.97 \\
Beauty     & 29.54 & 31.32 & 31.51 & 30.22 & 31.81 & 31.95 \\
Bosphorus  & 28.00 & 29.43 & 29.41 & 28.40 & 30.03 & 30.15 \\
Jockey     & 25.95 & 26.66 & 26.84 & 27.27 & 28.29 & 28.30 \\
ShakeNDry  & 25.70 & 29.18 & 29.27 & 27.84 & 30.15 & 30.33 \\
\hline
Average    & 26.10 & 28.15 & 28.37 & 27.41 & 29.33 & 29.50 \\
\hline
\end{tabular}}
\end{table*}

\begin{figure*}[!t]
\centering
\includegraphics[width=\textwidth]{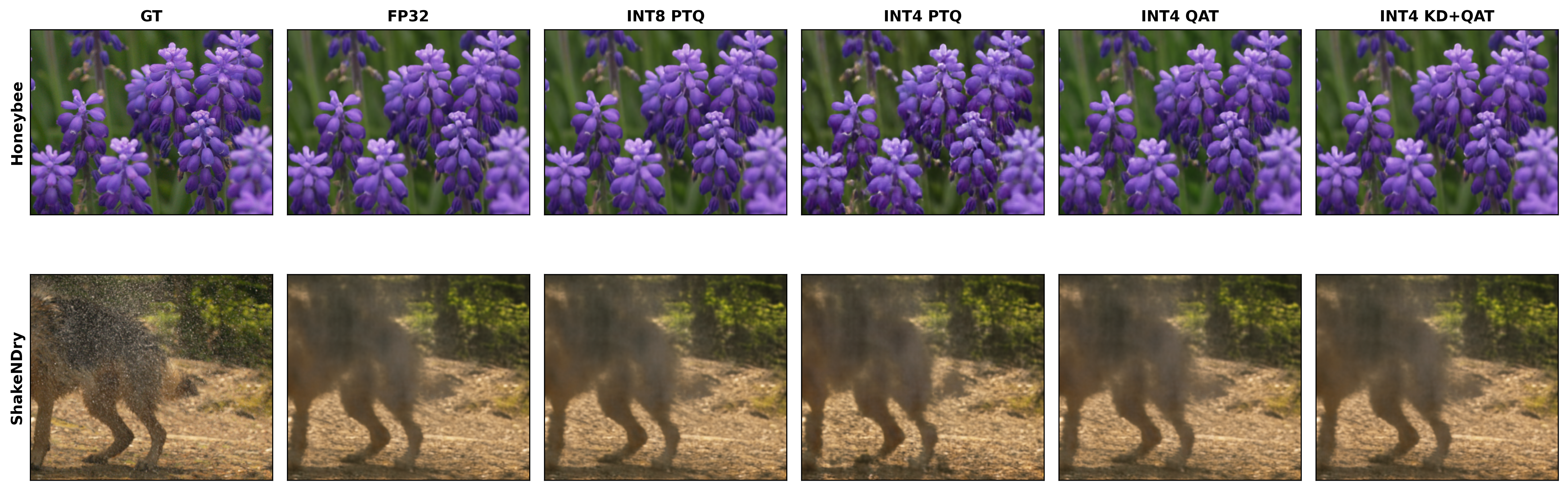}
\caption{\rev{Qualitative comparison of low-precision NeRV-T reconstruction on HoneyBee and ShakeNDry. Columns show ground truth (GT), FP32, INT8 PTQ, INT4 PTQ, INT4 QAT, and INT4 KD+QAT. INT8 remains close to FP32, while QAT and KD-assisted QAT recover much of the detail lost under INT4 PTQ.}}
\label{fig:int4_qualitative}
\end{figure*}

\rev{To complement these quantitative results, Fig.~\ref{fig:int4_qualitative} compares representative NeRV-T reconstructions from HoneyBee and ShakeNDry. INT8 PTQ remains visually close to FP32, whereas INT4 PTQ exhibits a clearer loss of local sharpness and fine structure. QAT substantially restores this structure, and KD+QAT provides a further improvement on both illustrated sequences.}

\rev{Overall, the complete UVG benchmark confirms the BBB trends under controlled cross-sequence conditions: NeRV-T+ is consistently higher fidelity, frequency--focal KD improves every sequence, INT8 remains near-lossless, and QAT becomes increasingly important as precision decreases. The smaller and content-dependent KD+QAT gain at INT4 further shows that capacity, precision, and teacher supervision interact with video content.}

\subsection{Discussion}

\rev{The results establish the tiny NeRV regime as a meaningful deployment region rather than merely an extreme ablation. NeRV-T prioritizes minimum parameter count and computation, while NeRV-T+ allocates additional width for consistently higher spatial and temporal fidelity. Section~IV-B further shows that these reductions correspond to concrete stored-weight and compute thresholds on representative embedded and edge-AI platforms, making the two variants complementary operating points for different resource budgets.}

\rev{Training-time supervision partially compensates for the representational limitations introduced by this scaling. Extended training alone provides comparatively small gains, whereas knowledge distillation improves reconstruction without increasing inference-time complexity. Prior frequency-aware distillation studies similarly show that compact students benefit from selectively transferring spectral information that is difficult to reproduce under limited capacity~\cite{zhang2022waveletKD,pham2024frequencyKD}. In TinyNeRV, the component ablation shows that both frequency decomposition and focal disagreement weighting improve upon conventional final-output distillation, while their combination performs best. This supports the intended interpretation of frequency--focal distillation: structural and detail-oriented supervision are separated, and greater emphasis is placed on regions where the student deviates most from the teacher.}

\rev{The complete UVG evaluation also provides evidence that the observed TinyNeRV behavior is not specific to a single sequence or content type. All seven UVG sequences and BBB are evaluated under the same 132-frame, $720\times1280$ protocol without sequence-specific tuning, yet the principal ordering of the capacity, distillation, temporal, and precision results remains consistent across heterogeneous content. NeRV-T+ provides higher reconstruction fidelity and temporal stability than NeRV-T throughout the benchmark, frequency--focal distillation improves both tiny variants on every UVG sequence, and increasingly aggressive quantization benefits more strongly from retraining-aware adaptation. Although the magnitude of these effects varies with video content, their overall behavior is preserved, supporting the scalability of the capacity--supervision--precision framework across different video sequences.}

\rev{The broader distillation principle may also extend beyond the present NeRV setting. Recent work in tabular anomaly detection shows that compact students do not preserve all forms of teacher knowledge equally, with global behavior transferring more reliably than local or neighborhood-dependent structure~\cite{krcmar2026tabularKD}. TinyNeRV exhibits an analogous visual effect: coarse structure is comparatively easier to retain, whereas fine spatial detail is disproportionately affected by width reduction. This suggests a more general principle of selectively emphasizing information that capacity-constrained students preserve poorly. The specific Gaussian low/high-frequency decomposition used here is naturally suited to spatial signals and could therefore be investigated in other dense reconstruction tasks such as image restoration or super-resolution. For non-spatial domains, however, the Gaussian formulation itself is not directly transferable; the underlying selective-supervision principle would require a domain-appropriate decomposition or difficulty measure.}

\rev{Numerical precision introduces a complementary deployment axis. Consistent with established quantization literature, moderate precision is comparatively robust while increasingly aggressive bit reduction benefits more strongly from retraining-aware adaptation~\cite{nagel2021quantization}. In TinyNeRV, INT8 remains close to the full-precision operating point, INT6 forms a transitional regime, and INT4 suffers substantial degradation under PTQ. QAT recovers much of this loss, while teacher supervision provides a further average improvement at INT4 and improves or effectively matches QAT in nearly all sequence--model comparisons. These results indicate that architectural capacity and numerical precision interact: when representational redundancy is already limited, coarse weight discretization becomes more difficult for the student to absorb.}

\rev{These findings lead to several practical operating guidelines. NeRV-T is most appropriate when minimizing conventional computation and stored-weight footprint is the primary objective, whereas NeRV-T+ provides a stronger quality--efficiency compromise when additional capacity is available. INT8 offers the safest reduced-precision setting across the evaluated content, while INT4 should be paired with QAT and teacher supervision when maximizing weight-storage reduction is the dominant requirement. The quantization results should nevertheless be interpreted as simulated weights-only experiments: activations remain in full precision and low-bit weights are dequantized for floating-point computation. Accordingly, the reported results characterize reconstruction sensitivity and stored-weight precision rather than native integer throughput, energy consumption, or accelerator-level latency.}

\rev{Several limitations remain. Although the evaluation covers the complete UVG benchmark plus BBB under a common protocol, all experiments retain the canonical single-video NeRV setting, with one model optimized per sequence. Generalization to multi-video representations, streaming segments, substantially different neural video architectures, and non-video reconstruction tasks therefore remains to be validated explicitly. Native low-bit execution is also not evaluated, so future work should measure latency, energy, and end-to-end memory behavior on embedded accelerators. The characterized capacity and precision points could additionally support adaptive systems that select model size or precision according to available compute, memory, and quality requirements. Overall, TinyNeRV provides a deployment-oriented basis for selecting compact neural video representations under explicit resource constraints.}

\section{Conclusion}

This work presented a systematic study of tiny Neural Video Representation models tailored for constrained deployment. We introduced two lightweight configurations, NeRV-T and NeRV-T+, and analyzed their efficiency--quality trade-offs under architectural scaling, knowledge distillation, and low-precision inference.

\rev{The results demonstrate that the low-capacity regime remains practically viable when supported by appropriate training and numerical adaptation strategies. Across the complete UVG benchmark, frequency--focal distillation improves both tiny variants on every sequence, while INT8 weights remain nearly lossless and INT6 degradation is largely recovered by QAT. At INT4, QAT provides the dominant recovery and teacher supervision yields a further positive average gain. These findings reveal a structured interaction between model capacity, supervisory strength, and numerical precision in compact neural decoders.}

\rev{Across all seven UVG sequences plus BBB, the principal capacity and precision trends remain consistent, strengthening the evidence that the tiny regime occupies a meaningful region of the efficiency--quality spectrum. NeRV-T offers the most aggressive compute-oriented operating point, while NeRV-T+ provides a higher-fidelity alternative under a larger but still reduced budget. Overall, this study establishes a deployment-oriented basis for selecting compact neural video representations under explicit resource and reconstruction-quality constraints.}

\section*{Data Availability}
The datasets used in this study are publicly available. The implementation is available at: \url{https://github.com/HannanAkhtar/TinyNeRV-Implementation}.

\section*{Acknowledgment}
Research reported in this paper was supported by a faculty startup grant from the American University of Sharjah (grant number: SE2605).
This article represents the opinions of the author and does not mean to represent the position or opinions of the American University of Sharjah.

\bibliographystyle{IEEEtran}
\bibliography{main}

\raggedbottom
\setlength{\parskip}{0pt}


\begin{IEEEbiography}[
{\includegraphics[width=1in,height=1.25in,clip,keepaspectratio]{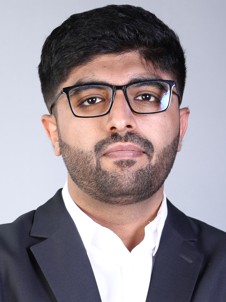}}
]{Muhammad Hannan Akhtar}
(Graduate Student Member, IEEE) received the M.Sc. degree in Computer Engineering from the American University of Sharjah (AUS), United Arab Emirates, in 2025, and the B.Sc. degree in Computer Engineering from Ajman University, United Arab Emirates, in 2021.

His research interests include machine learning, computer vision, neural video representations, efficient deep learning, video anomaly detection, reinforcement learning, and deployment-oriented AI for resource-constrained environments. His recent work focuses on compact neural video representations, low-precision inference, knowledge distillation, scalable model precision, and efficient reconstruction-based video anomaly detection. He has also worked on mobile-friendly deep learning models for agricultural insect classification and adversarial vulnerability assessment in Wi-Fi fingerprint localization systems.
\end{IEEEbiography}


\begin{IEEEbiography}[
{\includegraphics[width=1in,height=1.25in,clip,keepaspectratio]{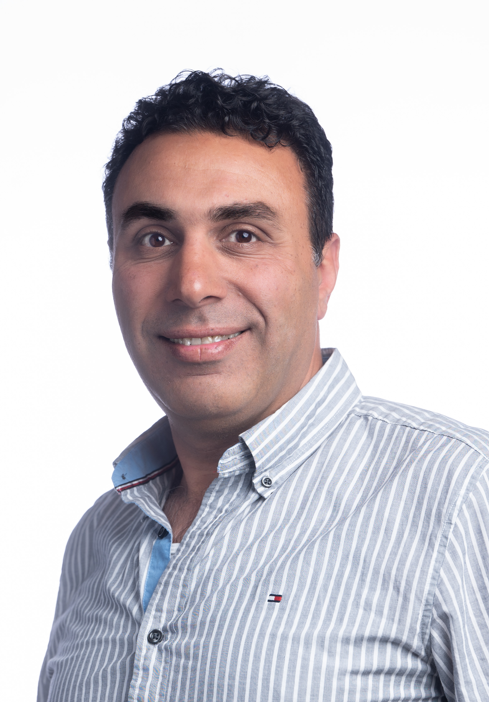}}
]{Ihab Amer}
(Senior Member, IEEE) received the Ph.D. degree in Electrical and Computer Engineering from the University of Calgary, AB, Canada, the M.Sc. degree in Computer Science from the American University in Cairo, Egypt, and the B.Sc. degree in Computers and Systems Engineering from Ain Shams University, Egypt. Since August 2025, he has been a Professor of Practice with the Department of Computer Science and Engineering at the American University of Sharjah (AUS), United Arab Emirates.

He previously held senior technical leadership roles in industry, including Fellow and member of the Software Technologies and Architecture Team at Advanced Micro Devices (AMD), Markham, Canada, where he led advanced initiatives in multimedia and AI architectures and chaired the multimedia and display patenting committee. Earlier, he served as an Assistant Professor at the German University in Cairo and as a consulting researcher with the ATIPS Labs, University of Calgary, with research visits to EPFL and Xilinx.

Dr. Amer's work spans learned and classical multimedia systems, video compression, AI and multimedia architectures, efficient computer arithmetic, and fault-tolerant and approximate computing. He has over 40 peer-reviewed publications, more than 40 contributions to MPEG standards, and over 50 issued and pending patents. He received several distinctions, including the AMD Innovation Award and ISO/IEC recognition for special contribution to MPEG-4 Part 9. He delivered a keynote at ACM MMSys 2023 and tutorials at ISCAS 2016 and ICECS 2015. He is a Senior Member of IEEE and a licensed Professional Engineer in Ontario, Canada.
\end{IEEEbiography}


\begin{IEEEbiography}[
{\includegraphics[width=1in,height=1.25in,clip,keepaspectratio]{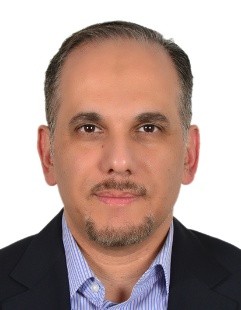}}
]{Tamer Shanableh}
(Senior Member, IEEE) was born in Scotland, U.K. He received the M.Sc. degree in Software Engineering in 1998 and the Ph.D. degree in Electronic Systems Engineering in 2002, both from the University of Essex, U.K.

He is currently a Professor of Computer Science with the American University of Sharjah, United Arab Emirates, which he joined in 2002. From 1999 to 2002, he served as a Senior Research Officer with the University of Essex, where he collaborated with BTexact on the invention of video transcoders. He subsequently joined Motorola U.K. Research Labs, where he contributed to the establishment of the Error Resilient Simple Scalable Profile within the ISO/IEC MPEG-4 standard.

Throughout his career, Dr. Shanableh has held several visiting professorships at Motorola Labs and completed a sabbatical leave as a Visiting Academic with the Multimedia and Computer Vision Lab at Queen Mary University of London, U.K. He holds six patents and has authored more than 100 publications, including 10 IEEE Transactions papers. His primary research interests include deep learning and video coding and processing.
\end{IEEEbiography}

\end{document}